\documentclass[review]{elsarticle}

\usepackage{lineno, hyperref}
\modulolinenumbers[5]

\usepackage{xcolor,soul,framed} 
\colorlet{shadecolor}{yellow}

\usepackage[cmex10]{amsmath}
\usepackage{array}
\usepackage{mdwmath}
\usepackage{mdwtab}
\usepackage{eqparbox}
\usepackage{url}

\usepackage{subfigure}
\usepackage{array}

\usepackage{changepage}
\newcommand{\tabincell}[2]{\begin{tabular}{@{}#1@{}}#2\end{tabular}} 
\newcommand\tab[1][1.5cm]{\hspace*{#1}}

\journal{Neurocomputing}

\begin{document}

\begin{frontmatter}

\title{Predicting Flight Delay with Spatio-Temporal Trajectory Convolutional Network and Airport Situational Awareness Map}

\author[rmit]{Wei Shao\corref{mycorrespondingauthor}\textsuperscript{\ddag}}
\cortext[mycorrespondingauthor]{Corresponding author}
\cortext[eqc]{\ddag Equal contribution}
\author[rmit,d61]{Arian Prabowo\textsuperscript{\ddag}}
\author[rmit]{Sichen Zhao}
\author[d61,anu]{Piotr Koniusz}
\author[rmit]{Flora D. Salim}

\address[rmit]{Royal Melbourne Institute of Technology, Melbourne, Australia}
\address[d61]{DATA61 / CSIRO, Australia}
\address[anu]{Australian National University, Canberra, Australia}





\begin{abstract}
To model and forecast flight delays accurately, it is crucial to harness various vehicle trajectory and contextual sensor data on airport tarmac areas. These heterogeneous sensor data, if modelled correctly, can be used to generate a situational awareness map. Existing techniques apply traditional supervised learning methods onto historical data, contextual features and route information among different airports to predict flight delay are inaccurate and only predict arrival delay but not departure delay, which is essential to airlines. In this paper, we propose a vision-based solution to achieve a high forecasting accuracy, applicable to the airport. Our solution leverages a snapshot of the airport situational awareness map, which contains various trajectories of aircraft and contextual features such as weather and airline schedules. We propose an end-to-end deep learning architecture, TrajCNN, which captures both the spatial and temporal information from the situational awareness map. Additionally, we reveal that the situational awareness map of the airport has a vital impact on estimating flight departure delay. Our proposed framework obtained a good result (around 18 minutes error) for predicting flight departure delay at Los Angeles International Airport.
\end{abstract}


\end{frontmatter}


\section{Introduction}
Flight delay has a significant negative economic impact, as well as being detrimental to the climate and international communications. In the United States alone, more than US\$26.6 billion are wasted due to the flight delays in 2017 based on the estimation of the Federal Aviation Administration (FAA) \cite{GAO19}. Flight delay and cancellation also takes responsibility for nearly a third of complaints from air travel passengers for selected airlines \cite{GAO19}. Additionally, around a quarter of all commercial flights have been delayed or cancelled \cite{GAO11} in recent years.

Departure delay, the most common and intolerable flight delay, caused more than 1600 overstayed flights in the airport for more than three hours during the summertime between 2004 to 2010 in the United States \cite{GAO11}. However, its root causes do not attract sufficient attention from researchers. Most existing works focus on other types of delays since departure delay is influenced by the air travel both and on-ground tarmac situations.

\begin{figure}
\includegraphics[width=\linewidth]{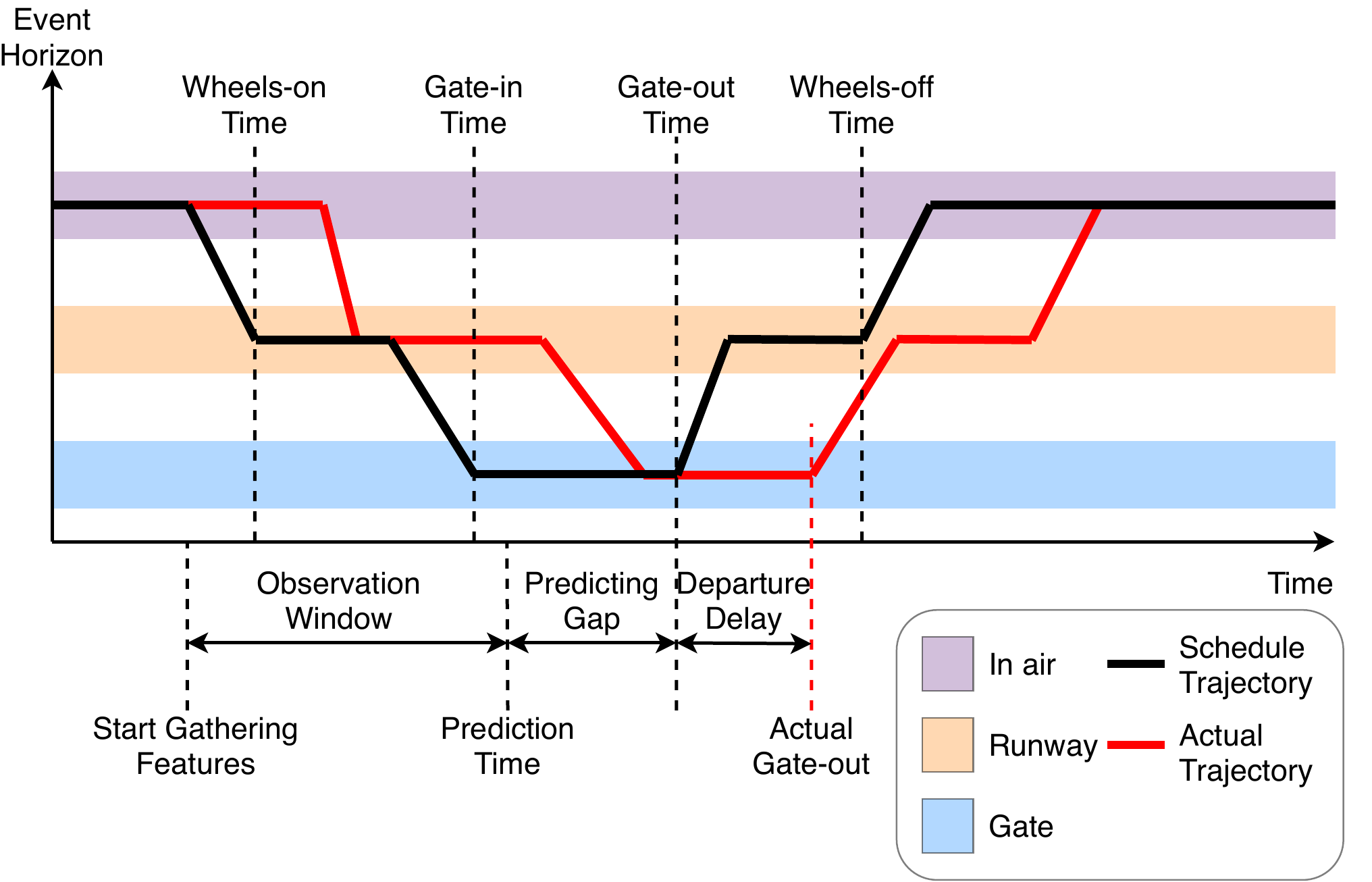}
\caption{An illustration of flight delay. We predict the departure delay which is between the real gate-out time and scheduled gate-out time using the data extracted from observation window.} \label{fig_flight}
\end{figure}

Figure \ref{fig_flight} illustrates the concept of departure delay by showing the scheduled against the real \textit{flight travel flow}. The \textit{flight travel flow} usually comprises the following five stages: the airborne stage, taxi-in stage, turn around stage, taxi-out stage and airborne again. \textit{Departure delay} refers to the period between the gate departure time and gate arrival time. The departure delay causes passenger anxiety and the uncertainty of scheduled economic and social activities. Therefore, an accurate departure delay prediction model is needed to alleviate these problems \cite{GAO19}.

Most aerospace experts explored the correlation between environmental issues (weather, wind, etc.), attributes of flight data (day of a week, season, month, etc.) and flight delay time \cite{Kim2016}\cite{Sternberg2017}. There are also recent works in relevant Computer Science fields that have explored probabilistic models \cite{ahmadbeygi2010decreasing, tu2008estimating}, network representation \cite{abdelghany2004model, tu2008estimating}, and machine learning models \cite{rebollo2012characterization, ahmadbeygi2008analysis}. However, the problem of departure delay is under-studied and more complex, as it is influenced by the environmental factors, air traffic and also on-ground airport situations. 

Aerospace expert researchers have proposed a concept called air traffic complexity (ATC) to describe airport traffic \cite{mogford1995complexity}. However, most works rely on established equations based on empirical observations without offering any quantification of the concept. In this context, Internet of Thing (IoT) -related  techniques such as sensor networks and big data analytics have the potential to provide a scalable solution. Increasing works are exploring sensor networks in the air traffic area \cite{trub2018monitoring, prabowo2019coltrane, shao2019onlineairtrajclus}. They either use the sensor data to estimate the traffic at the airport or monitor meteorological parameters.

In this work, we address whether it is possible to predict the flight departure delay using big data from sensor networks, harnessing machine learning and aerospace domain knowledge. Firstly, we represent the ATC using airport situational awareness maps (the trajectory of flights and vehicles at the airport). This work bridges the gap between the ATC -- a concept proposed by aerospace experts, sensor data fusion, and machine learning techniques. We apply two ways to represent the ATC: 1) we propose a group of features which are extracted from sensor and radar data of aircraft at the airport, and 2) regard the whole airport as a grid map and take a snapshot of each time period. The locations, speed and other contextual information are represented as a sequence of images. Secondly, we applied the factor analysis to both environmental data and ATC data to explore the correlation between flight departure delay and situational awareness map (ATC, weather, etc.). Thirdly, both of traditional machine learning methods and a deep learning framework are used to predict the flight departure delay with a real-world dataset of the Los Angeles International Airport (LAX).

The experimental results show that both the features we proposed to represent the ATC and the sequence of images we used to represent the airport situational awareness map can be used to predict the flight departure delay, and they are more effective than weather conditions and schedules. Additionally, our proposed end-to-end deep learning framework also can predict the flight departure delay with the comparable performance with machine learning-based methods. We show that both the hand-crafted features based methods and the end-to-end deep learning approaches can achieve similar results.

The main contributions of this paper are the following:
\begin{itemize}
\item We propose a generic framework to predict the departure delay using situational awareness maps, original flight schedules and weather conditions.
\item We borrow a concept from aerospace area called Air Traffic Complexity (ATC), and use both aircraft trajectories and a sequence of snapshot of airport aircraft to represent this concept. We also analyse the importance of those features and compare them against previous delay-causing factors such as weather conditions and schedules.
\item We propose an approach to project the situational awareness map to a sequence of images and propose an end-to-end deep learning framework to predict the departure delay. The experimental results demonstrate the effectiveness of the approach to achieve a good result (less than 20 minutes prediction error).  
\end{itemize}

The paper is organised as follows: Section \ref{relatedwork} discusses the related work; our proposed methods are shown in Section \ref{chapter:methodology}; Section \ref{chapter:ExperimentAndResults} shows the experiments results; Section \ref{chapter:Discussion} discuss the current and future work, and Section \ref{chapter:conclusion} concludes the paper.

\section{Related Work}\label{relatedwork}


There exists a large number of studies in the field of air traffic complexity (ATC). However, the literature review shows that the majority of them are considering using the complexity factors to indicate the currently existing or expected impact on air traffic controller's workload. Additionally, various studies on aircraft trajectory sensor data have shown to be effective in mobility forecasting and map creation \cite{vouros2018big, andrienko2018creating}.

A polynomial equation was proposed by \cite{mogford1995complexity} which described ATC as the combination of static sector characteristics and dynamic traffic patterns. Another metric approach called Dynamic Density (DD) was proposed as a measure of ATC by NASA \cite{laudeman1998dynamic} and quickly became a well-known metric. The DD metric is defined as a linear combination of multiple weighted traffic complexity factors (TC); traffic density (TD) and air traffic controller intent (CI). 

A study proposed by NASA in 2007 \cite{kopardekar2007airspace} shows that the performance of the complexity metrics varies at different facilities and conduct a further validation on the 52 complexity variables for the Cleveland Centre.
The paper by \cite{djokic2010air} lists of all complexity factors proposed by the predecessors is given and an attempt to reduce the size of that factor set is presented. The focus of the research by \cite{delahaye2003air} is to find a non-linear dynamical model which can differentiate the level of the air traffic complexity based on vectorised aircraft locations.

Since the researches mentioned above are mainly focused on the relationship between ATC and air traffic controller workload, which normally exists in the en-route environment. 
The research about airfield and airport traffic complexity has received very little attention (\cite{koros2003complexity}, \cite{simic2015airport}), while the relationship between the air traffic complexity and the gate-hold delay has had even fewer attention \cite{tu2008estimating}. The study by \cite{koros2003complexity} has a primary focus on the ATC around the tower area and its influence on the air traffic controllers' workload. It also shows that the relative effectiveness of complexity metrics is site-specific due to the different layouts and configurations of every airport. 
\cite{simic2015airport} proposed a new metric called Dynamic Complexity which contains the features about the layout of a certain airport and the traffic interactions in the airport itself and its vicinity.

When considering predicting the departure delay of flights, the air traffic complexity can be seen as one of the most important parts, since it connects and captures the characteristics of several problems: the arrival delay propagation problem, the runway sequencing problem. 
In \cite{rebollo2014characterization}, a snapshot of the current departure delay state is used to characterise the current network state. This snapshot is a 584-dimensional vector which comprises the current departure delay state of each link in a simplified US airport network and the delay predictions at time $t$ are made based on the snapshot at time $t$. The main focus of \cite{xu2008multifactor} is to classify the major factors that could cause or absorb the flight delay. \cite{tu2008estimating} discussed the seasonal weather trend the arrival flight delay propagation effect but neglected factors that may describe the air traffic complexity.
In \cite{tu2008estimating}, departure delay is measured by the discrepancy between the scheduled departure time and the actual departure time from the gate.

\section{Definitions}\label{chapter:definition}

\subsection{Reference}
In what follows, we define a single flight, $\mathbf{f} = (\mathbf{f}_{\mathbf{x}} , \mathbf{f}_{\mathbf{y}})$, as a pair of vectors. The vectors, $\mathbf{f}_{\mathbf{x}}$ and $\mathbf{f}_{\mathbf{y}}$, represent the feature and label vectors respectively. Both vectors contain ordered pairs of attributes and values associated with a single flight. All flights, $\mathbf{f}$, are contained in the set, $\mathbf{F}$. Furthermore, all flight features, $\mathbf{f}_{\mathbf{x}}$, and labels, $\mathbf{f}_{\mathbf{y}}$,  belong to their respective sets, $\mathbf{F}_{\mathbf{x}}$ and $\mathbf{F}_{\mathbf{y}}$. The attributes of a feature vector are listed in Table \ref{tab:feature_vectors} and Table \ref{tab:weather_vectors}. This is also referred as the reference data source.

\begin{table}[htbp]
\caption{The attributes feature vectors in the reference data source.}\label{tab:feature_vectors}
\centering
\begin{tabular}{l||l}
\hline 
\hline 
\textbf{Attribute Name} & \textbf{Description}\\
\hline
\tabincell{l}{Scheduled departure \\time} & \tabincell{l}{The scheduled time when the aircraft \\would depart from the gate. Also known \\as gate-out time.}\\
\hline
\tabincell{l}{Scheduled elapsed \\time (departure)} & \tabincell{l}{The scheduled duration when the aircraft \\is on the air, between the current airport \\(LAX), and the destination airport.}\\
\hline
\tabincell{l}{Scheduled arrival \\time} & \tabincell{l}{The scheduled time when the aircraft \\would arrive at the gate. Also known as \\scheduled gate-in time.}\\
\hline
Actual arrival time & \tabincell{l}{The actual time when the aircraft arrived \\at the gate. Also known as actual \\gate-in time.}  \\
\hline
\tabincell{l}{Scheduled elapsed \\time (arrival)} & \tabincell{l}{The scheduled duration when the aircraft \\is on the air, between the origin and the \\current airport (LAX).}\\
\hline
\tabincell{l}{Actual elapsed time \\(arrival)} & \tabincell{l}{The actual duration when the aircraft \\was on the air,
between the origin and the \\current airport (LAX).} \\
\hline
Wheel on Time & \tabincell{l}{The actual time when aircraft wheel \\touched the runway.} \\
\hline
Delay carrier (arrival) & \tabincell{l}{One of the five component of arrival delay.} \\
\hline
Delay weather (arrival) & \tabincell{l}{One of the five component of arrival delay.} \\
\hline
\tabincell{l}{Delay national aviation \\system (arrival)} & \tabincell{l}{One of the five component of arrival delay.} \\
\hline
Delay security (arrival) & \tabincell{l}{One of the five component of arrival delay.} \\
\hline
\tabincell{l}{Delay late aircraft \\(arrival)} & \tabincell{l}{One of the five component of arrival delay.} \\
\hline
Arrival delay & \tabincell{l}{The total arrival delay.} \\
\hline
\hline 
\end{tabular}
\end{table}


Meanwhile, the attributes of a label vector are:

\begin{enumerate}
	\item Delay Carrier (departure)
	\item Delay Weather (departure)
	\item Delay National Aviation System (departure)
	\item Delay Security (departure)
	\item Delay Late Aircraft Departure (departure)
	\item Departure Delay
\end{enumerate}

The Departure Delay is the sum of the total of all the other five types of delays. Therefore, the performance metric is only based on the Departure Delay attribute. The remaining attributes are used for labels to help with training.

\subsection{ATC Features}

We define a GPS point, $\mathbf{p} \in \mathbf{P}$, as a vector of ordered pairs of attributes and values. The attributes are:

\begin{enumerate}
	\item latitude
	\item longitude
	\item time
	\item speed
	\item heading
\end{enumerate}

There are also other attributes such as altitude and aircraft type, but we removed these attributes during the data pre-processing step \ref{sec_Data_cleaning}. These GPS points are the basis for engineered features to capture Air Traffic Complexity (ATC) as done in \cite{shao2019flight}. In this paper, we propose a novel set of ATC features called TrajCNN Features based on these GPS points, described in Section \ref{sec_trajcnn_features}.

\subsection{Problem definition}

Most airlines only have real-time details about their flights, but not flights from other airlines. For this reason, to predict the Departure Delay, we will only use features from the flight in question, $\mathbf{f}_\mathbf{x}$, and not from any other flights. However, to capture the spatio-temporal information on the airport tarmac, airports could produce real-time aggregated and de-identified information with posing significant privacy or security risk. We call these kinds of information as ATC, defined in the previous subsection.

Two factors affect the decision of when to make the prediction. On the one hand, the prediction should be made as late as possible in order to capture the most recent and relevant information. On the other hand, the earlier the prediction, the more useful it will be for the airlines and the passengers. For example, earlier delay predictions would allow passengers to adjust their schedules, while late delay predictions might not allow for passengers to make any adjustment, and thus, rendering the prediction obsolete. We decided that a realistic predicting gap is four hours as passengers are expected to arrive at the airport 3 hours before departure, giving appropriate time for passengers to adjust their schedule before they arrive at the airports. We call this duration predicting gap.

Our model will capture all the tarmac spatio-temporal information within an observation window by creating a subset of $\mathbf{P}$, called $\mathbf{P}'$. In turn, the ATC features are constructed from $\mathbf{P}'$. Figure \ref{fig_flight} illustrates predicting the time, observation window and the definition of departure delay, with is the difference between the scheduled and actual departure (gate-out) time.

Formally, for every flight, $\mathbf{f} \in \mathbf{F}$, we predict the Departure Delay, $\mathbf{f}_\mathbf{y}$, using only features about that particular flight, $\mathbf{f}_\mathbf{x}$, and ATC constructed from a subset of GPS points, $\mathbf{P}'$.


\section{Dataset Analysis}

\begin{figure*}[htbp]
\subfigure[Detailed statistics of historical flight delay in our dataset.]{
\includegraphics[width=0.49\textwidth]{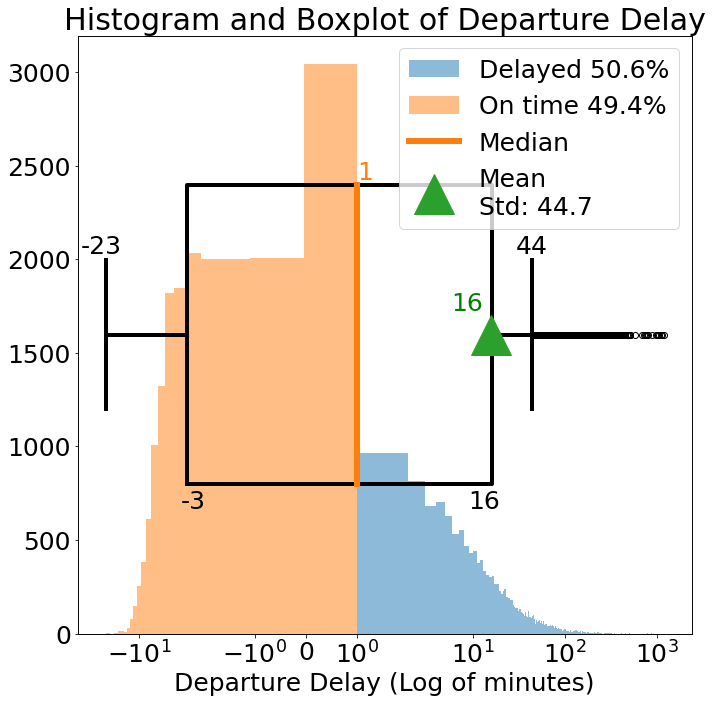}
\label{fig_dd}
}
\subfigure[Impact of Day of Week to flight delay.]{
\includegraphics[width=0.49\textwidth]{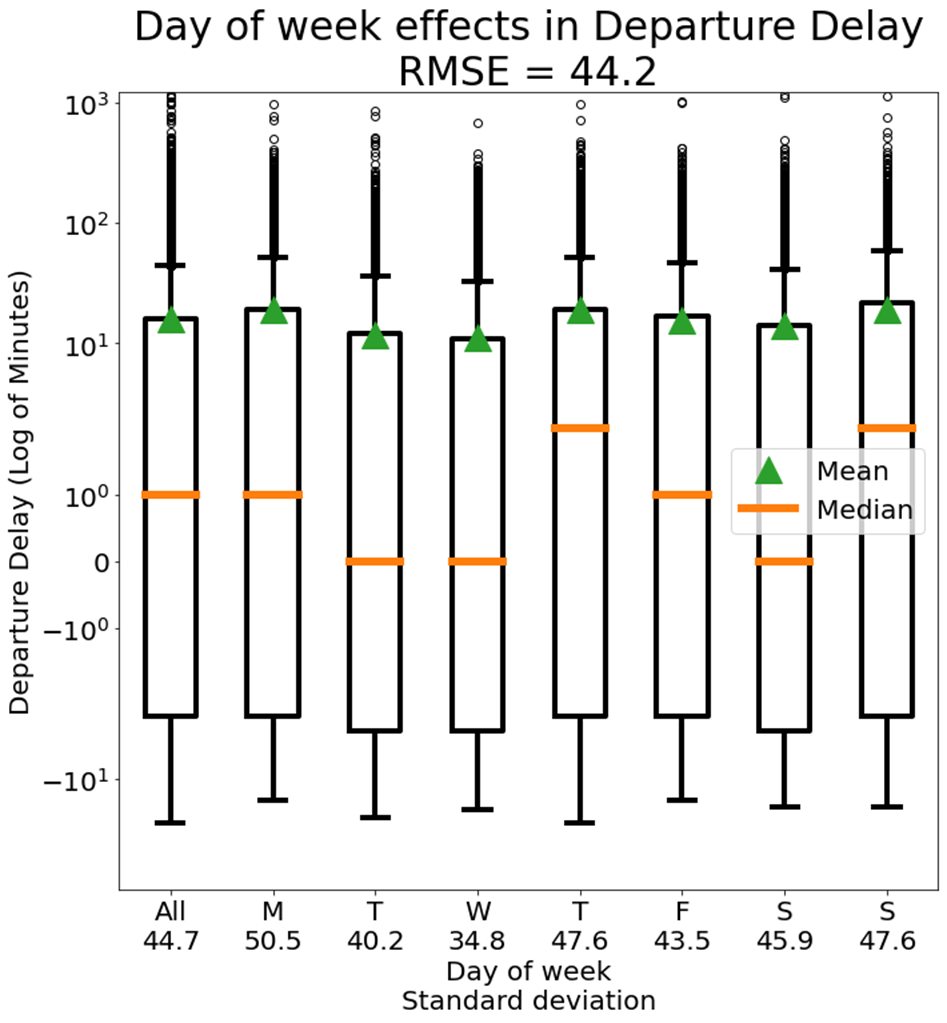}
\label{fig_dow}
}
\subfigure[Impact of Tine of Day to flight delay. There were no flights between 3 and 5 AM.]{
\includegraphics[width=\textwidth]{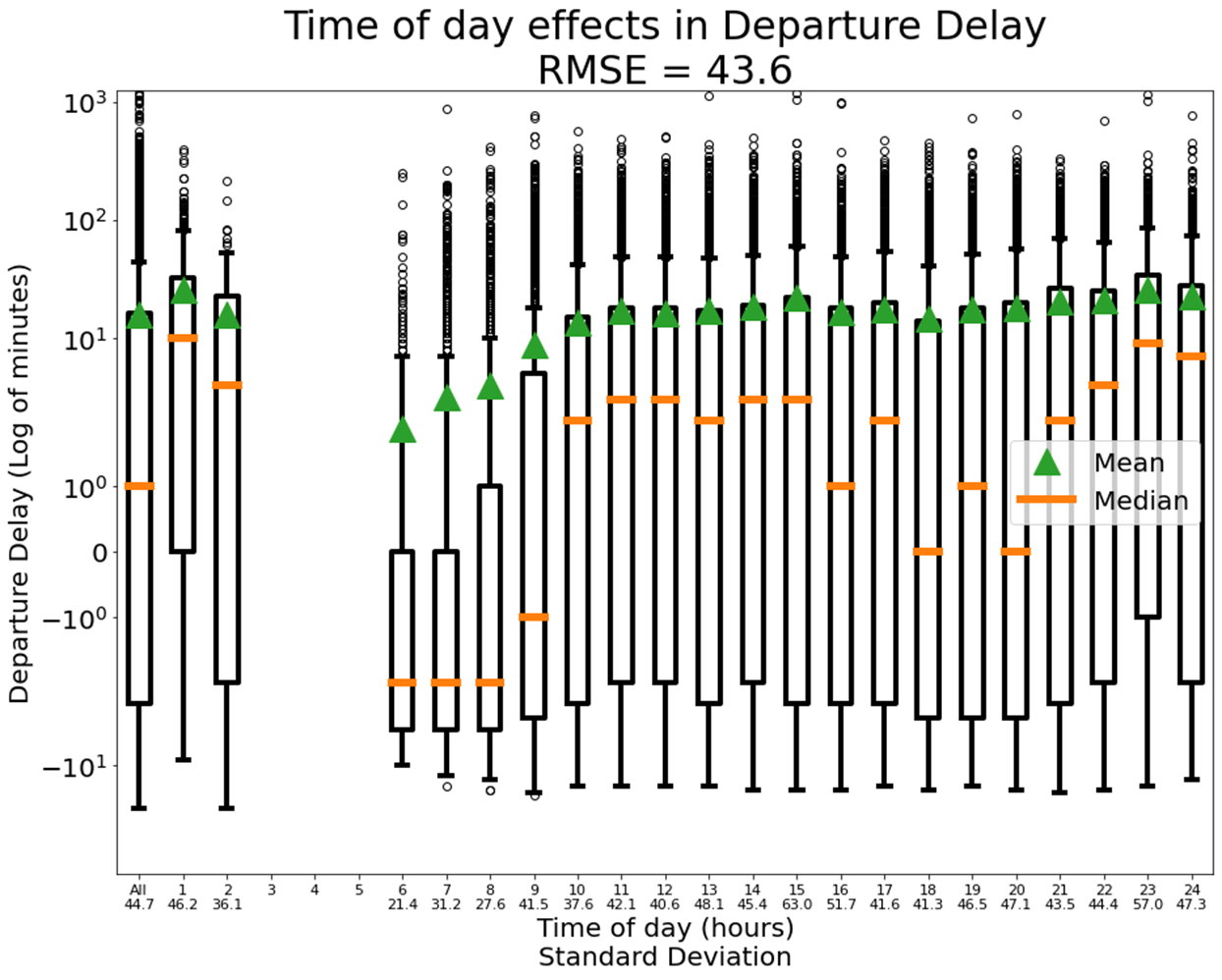}
\label{fig_tod}
}
\caption{Detailed statistics of historical flight delay and the impact of temporal features.}
\label{fig_dow_tod}
\end{figure*}

\subsection{Flight Delay}

Figure \ref{fig_dd} shows the detailed statistics of the flight delay in our dataset. Note that departure delay exist for all flights, not just the ones delayed. Also, departure delay can take negative value, meaning that the aircraft depart from the gate early. Using very strict definition of on-time: departure delay $\leq 0$, only a little bit less than half of the flights are on time (49.4\%) while the remaining are delayed by at least one minutes. This is consistent with the median of the delay being 1 minutes. However, this data is very skewed, as flights are unlikely to leave very early, yet very long delay is more likely, thus the mean of the delay (even including early departures) is 16 minutes, far higher than the median. The problem of flight delay prediction is made difficult not by the average magnitude of the delay, but by the high variability, measured by the standard deviation of 44.7 minutes.

It might be the case that these delay might easily be explained by a few factors. However, in the remaining of this section, we will show that flight delay prediction is a complex problem as it is not easily explained by any of the features alone. Simple combinations of the features are explored in subsection \ref{sec_exp_res}.

The explainability of a feature is calculated through RMSE. The RMSE is calculated by taking the average of each class and using it as the prediction. Note that if we consider the entire dataset as one single class, then the RMSE is equal to the standard deviation by definition. For this reason we use RMSE as comparison of delay factor between different features.

Note that we do not use the airlines and airports information as features in any of our models. The information are only provided here for completeness. The full list of features are available in Table \ref{tab:feature_vectors} and Table \ref{tab:weather_vectors}.

\subsection{Day of Week and Time of Day}

The first features we are going to analyse are the temporal features: Day of Week (Figure \ref{fig_dow}) and Time of Day (Figure \ref{fig_tod}). Although the mean and median delay fluctuates throughout the week, the standard deviations remained high. Taking the Day of Week effects into account, RMSE only goes down 0.05 from the standard deviation, showing that the Day of Week effect is small.

There is a stronger Time of Day effect compared to Day of Week effect. In the morning before 10 AM, when the airport is more quiet, there are much less delays. Nevertheless, the standard deviation within each hour is still very high. Although the RMSE from Time of Day is smaller than Day of Week, it is still only 1.1 minutes lower than the standard deviation.

\subsection{Airlines and Airports}

\begin{figure*}[htbp]
\centering
\subfigure[]{
\includegraphics[width=0.49\textwidth]{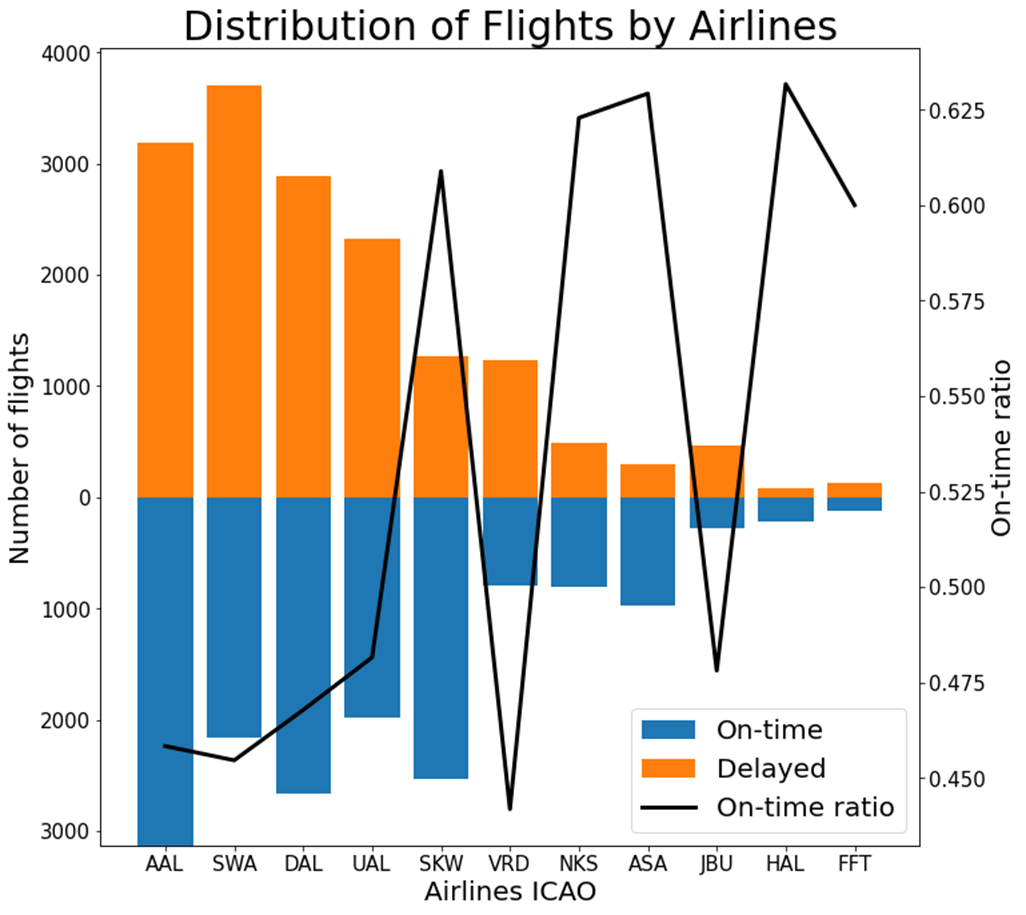}
\label{fig_al_hist}
}
\subfigure[]{
\includegraphics[width=0.41\textwidth]{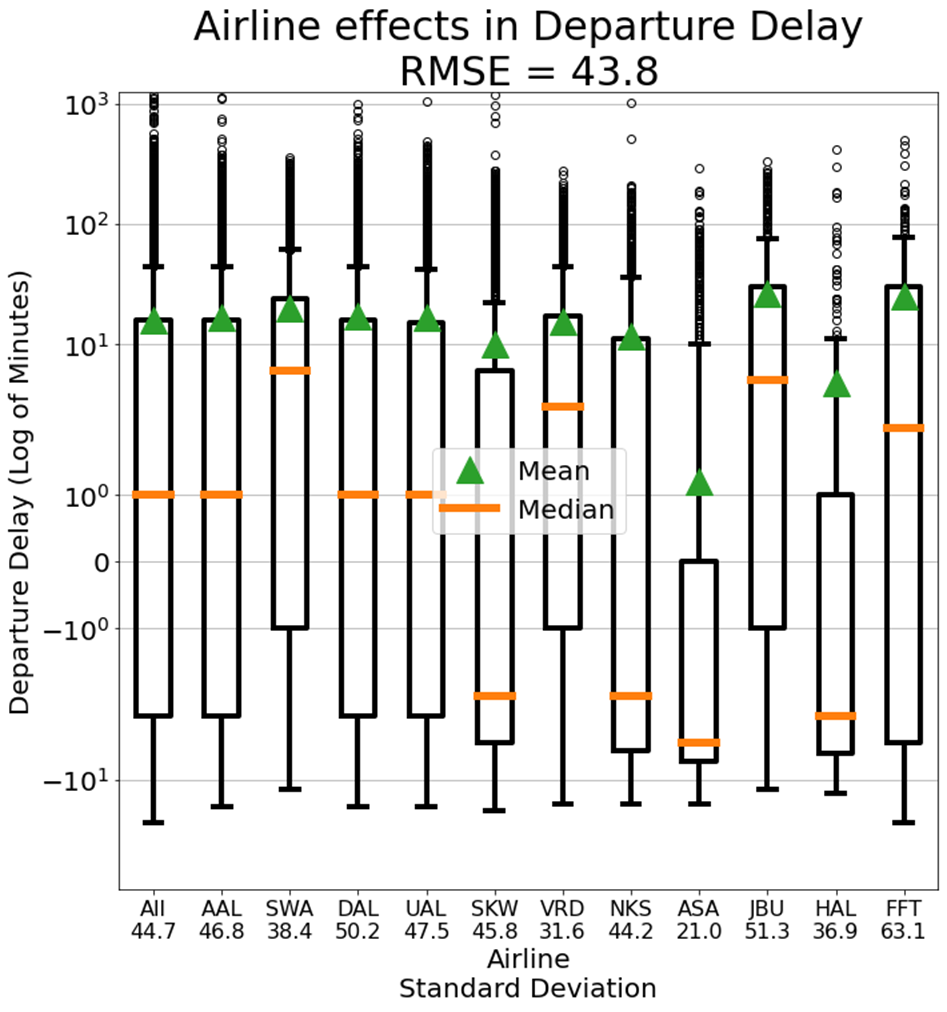}
\label{fig_al}
}
\subfigure[]{
\includegraphics[width=0.49\textwidth]{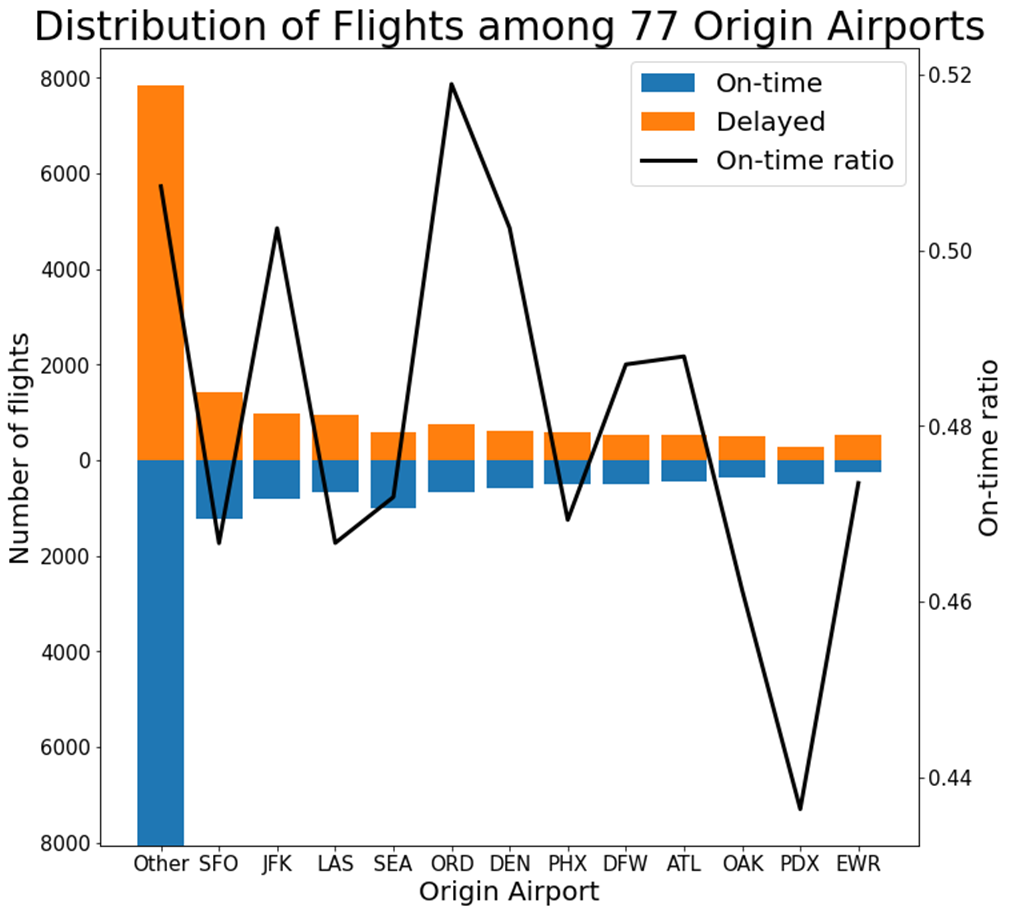}
\label{fig_ao_hist}
}
\subfigure[]{
\includegraphics[width=0.41\textwidth]{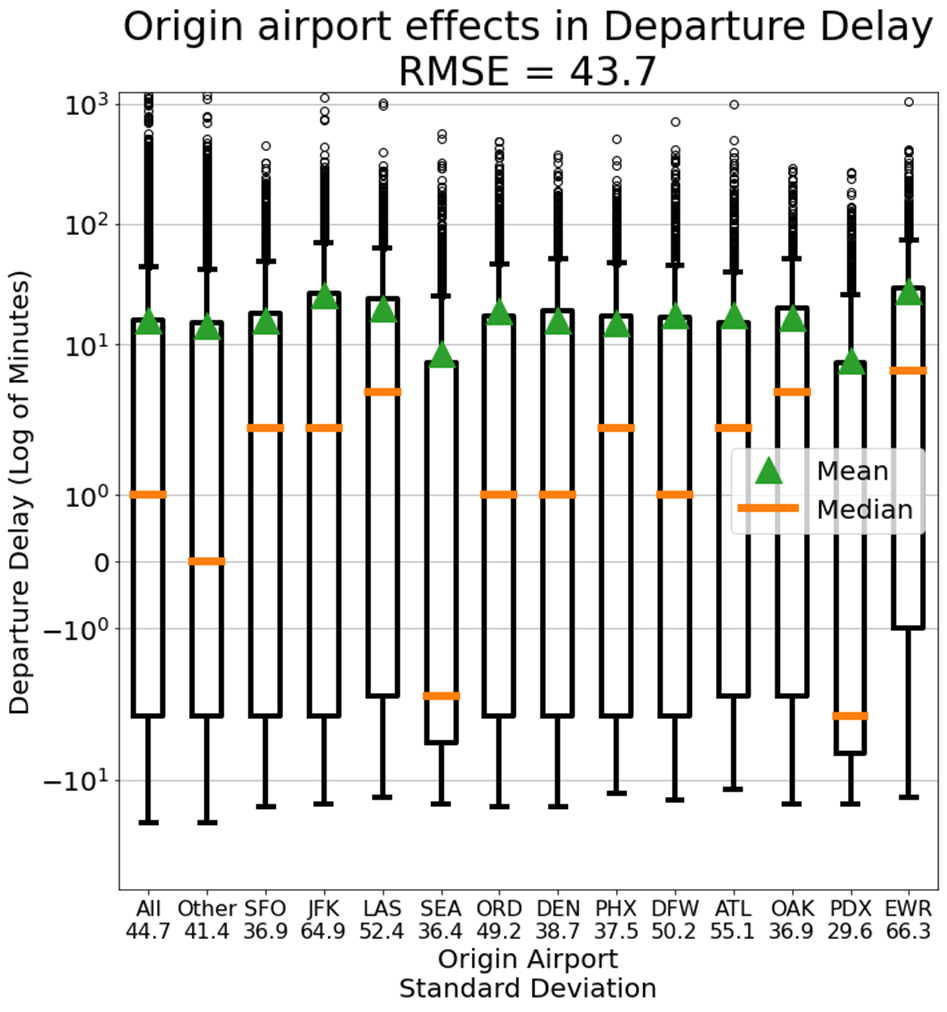}
\label{fig_ao}
}
\subfigure[]{
\includegraphics[width=0.49\textwidth]{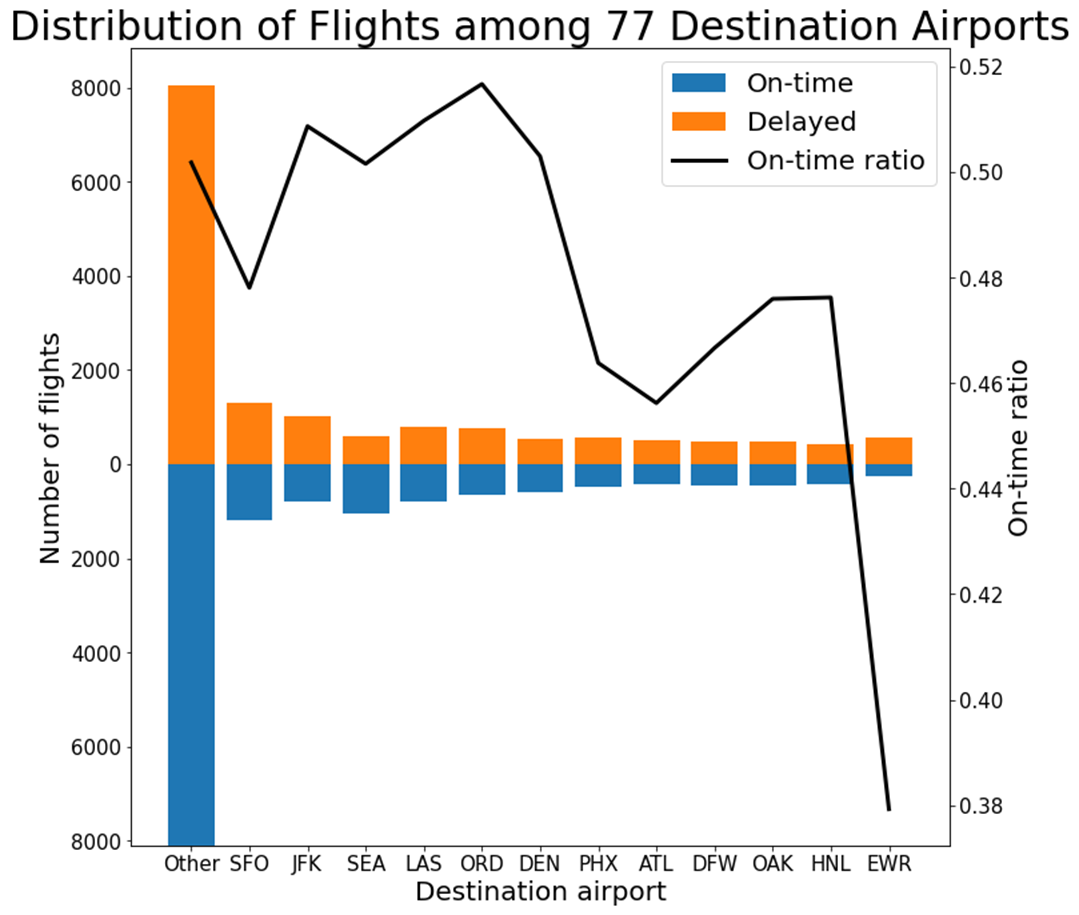}
\label{fig_ad_hist}
}
\subfigure[]{
\includegraphics[width=0.41\textwidth]{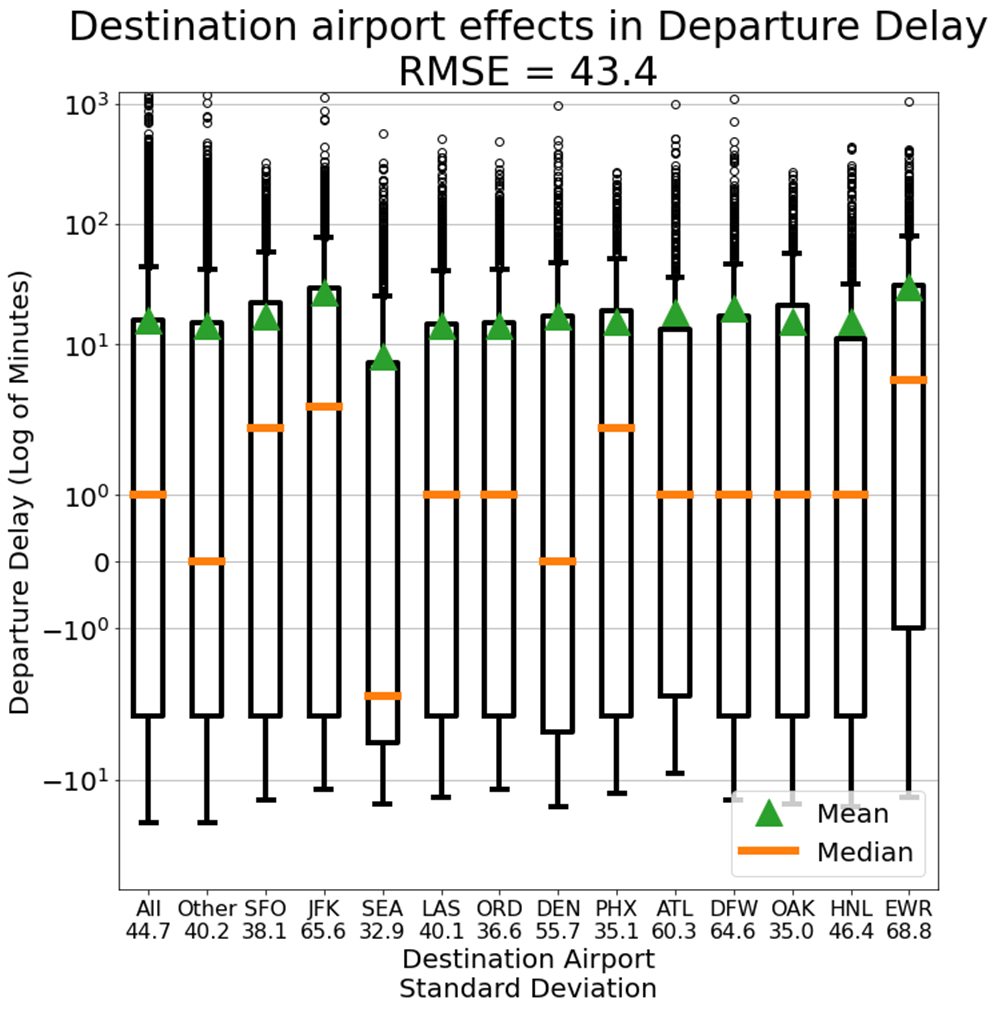}
\label{fig_ad}
}
\caption{The effect of airline and airport on departure delay.}
\label{fig_al_ap}
\end{figure*}

Next, we look at the effects of airlines, origin, and destination airports as shown in Figure \ref{fig_al_ap}. The common theme is that although some airlines and some airports has more delays than others, the variance within each group is still big, with the lowest being destination airport with RMSE of 43.4.

As there are 77 origin and destination airports, we only pick the top 12 as a part of our visualisation, and group the remaining into others. We pick the number 12 as the combinations of these 12 airports represent more than 50\% of all the flights.

\subsection{Weather}

\begin{figure*}[htbp]
\centering
\subfigure[]{
\includegraphics[width=0.9\textwidth]{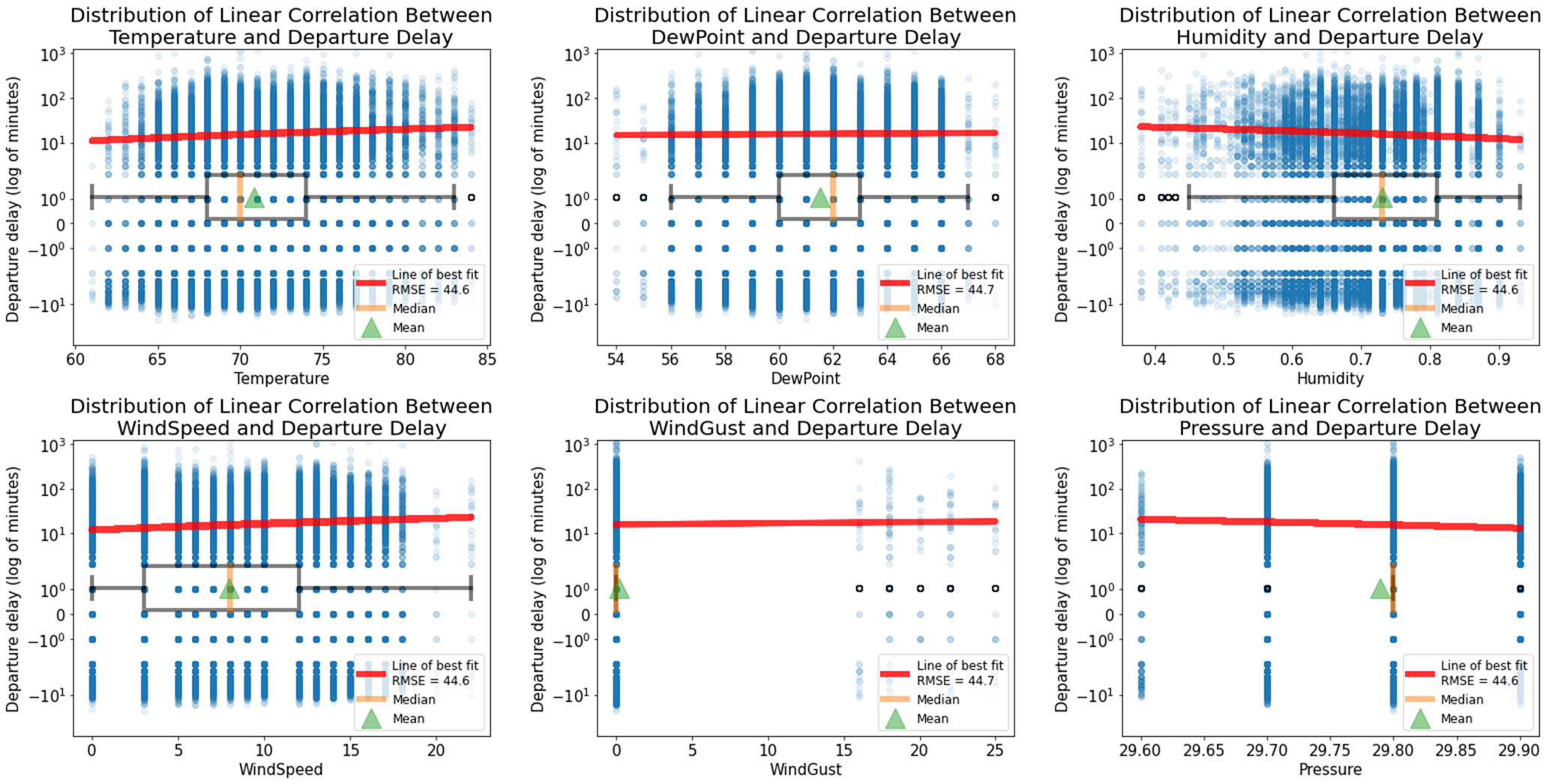}
}
\subfigure[]{
\includegraphics[width=0.46\textwidth]{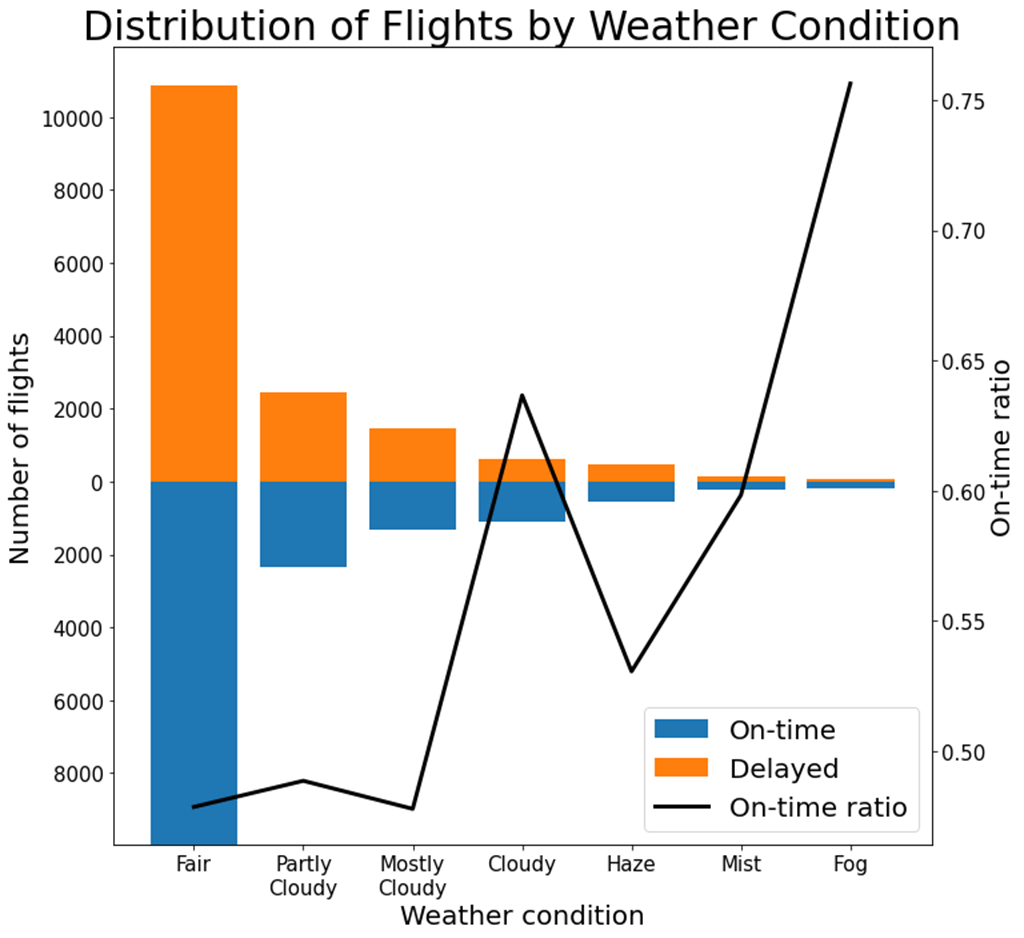}
}
\subfigure[]{
\includegraphics[width=0.41\textwidth]{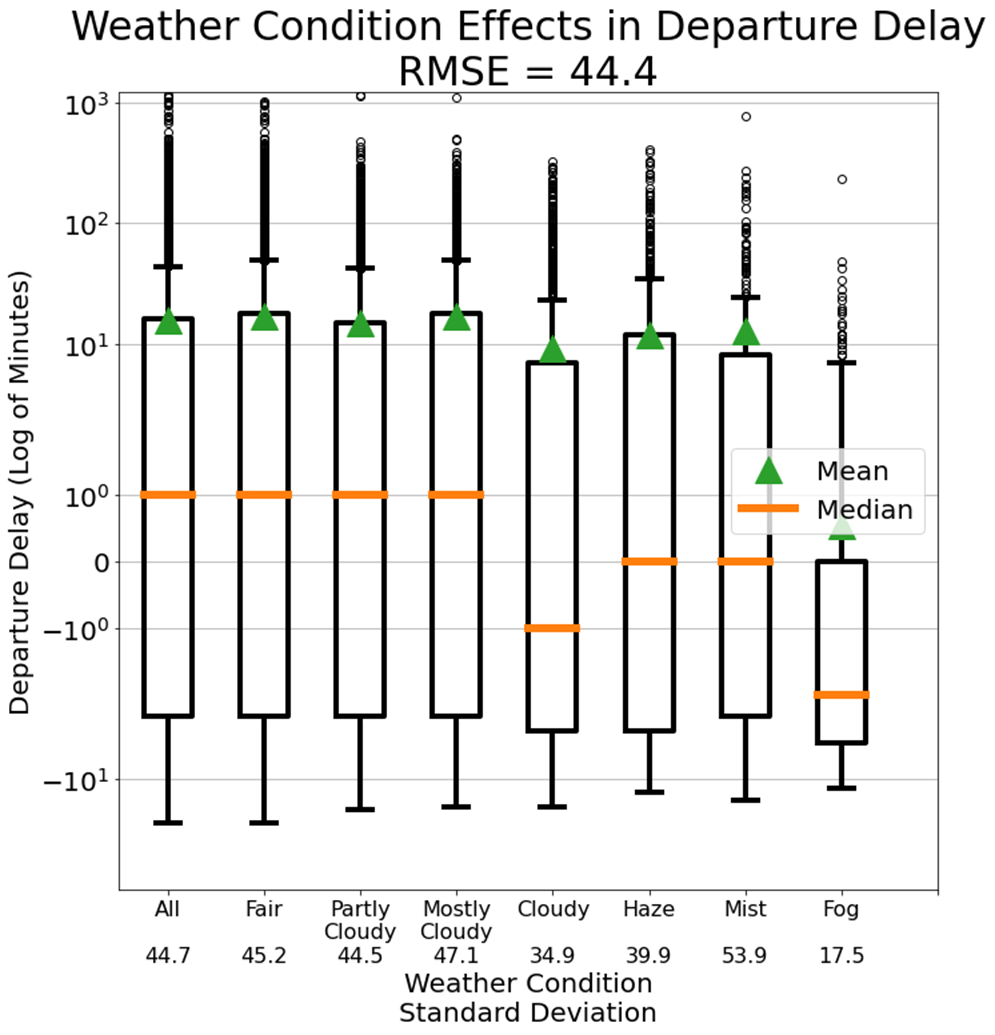}
}
\subfigure[]{
\includegraphics[width=0.46\textwidth]{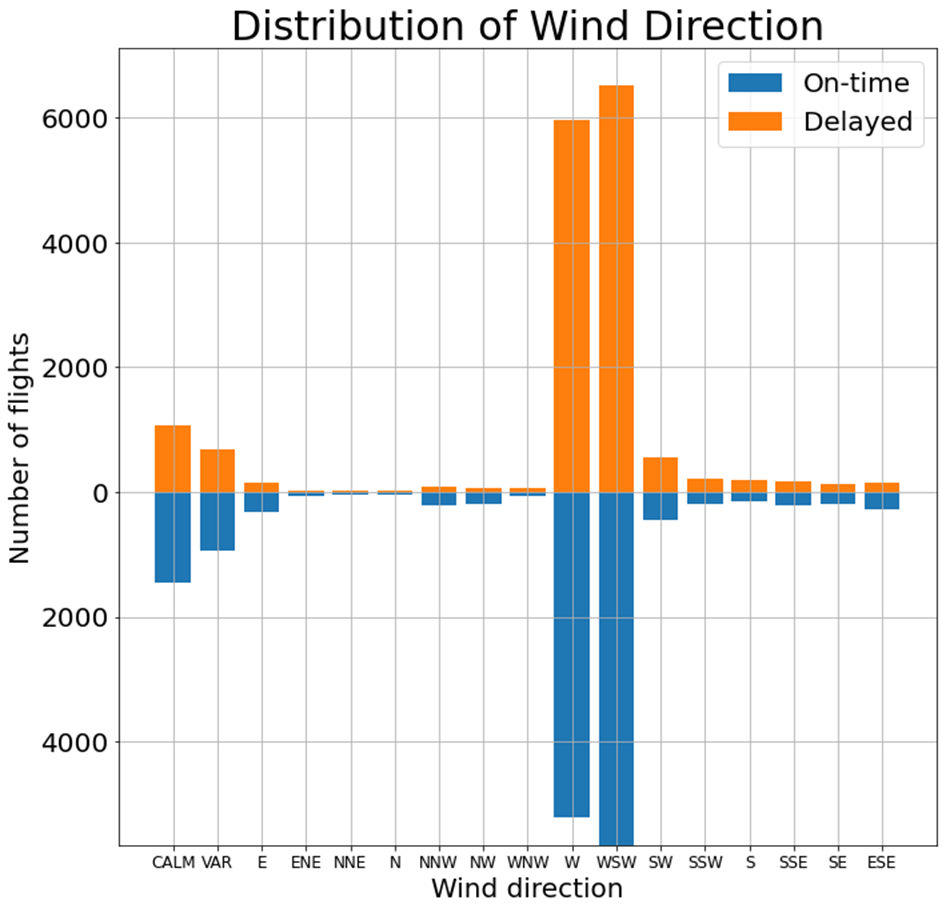}
}
\subfigure[]{
\includegraphics[width=0.41\textwidth]{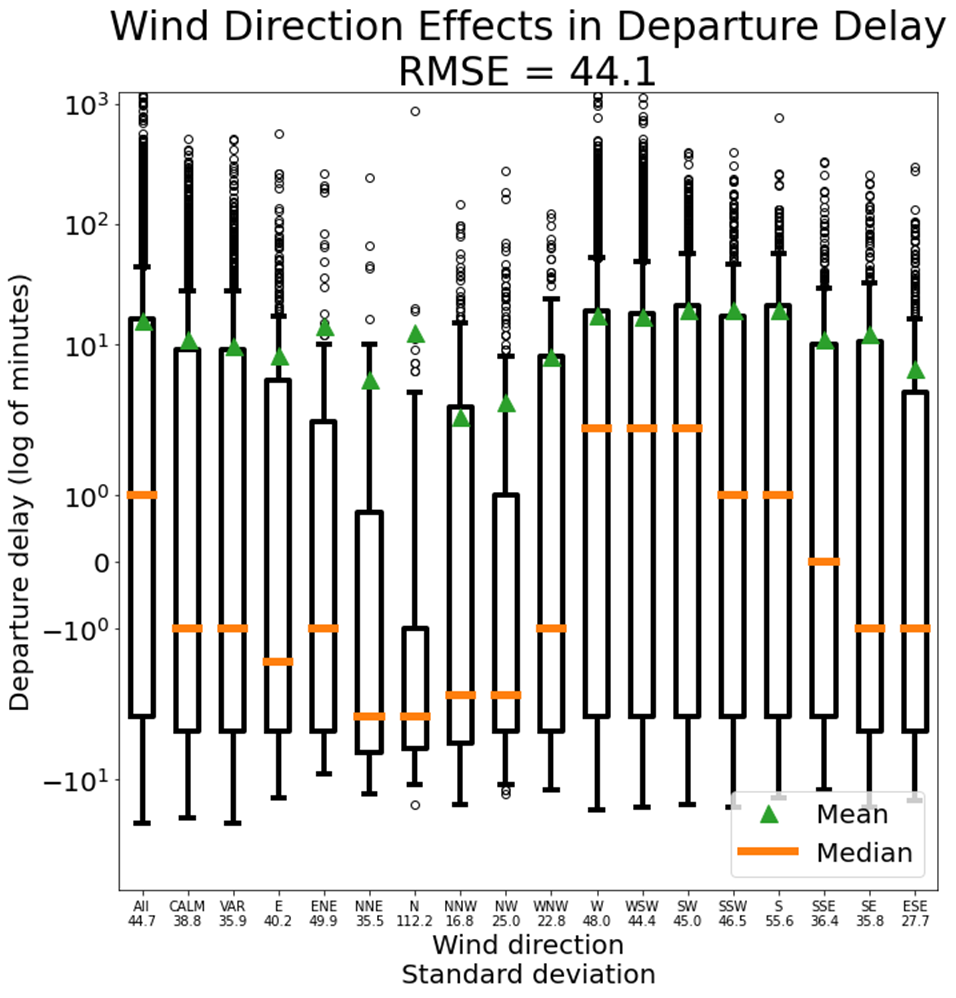}
}
\caption{The effect of weather on departure delay.}
\label{fig_wthr}
\end{figure*}

Finally, we analyse the relationship between many component of weather with flight delay in Figure \ref{fig_wthr}. For numerical variables, we draw the line of best fit and calculate the RMSE from those. The weather condition of most flight is fair, as expected from Californian cities. However, there are still sufficient variability in weather conditions, temperature, humidity, and wind speed. Contrary to expectation, fair weather has the lowest on-time ratio. This can be explained by the fact that the standard deviation of delay during fair weather is higher than the dataset (45.2 vs 44.7). This again suggested fair weather is not a significant factor regarding delay. As also expected from most places, the geography of a city is a huge determinant of wind direction, and thus we see that the wind direction as clustered around Westerly winds.

In this section, we conclude that temporal, airlines, airports, and weather factors alone cannot satisfactory explain the huge variability in airport delay. This suggests that we might need more information, such as spatial information from within the airport, or that there are complex interplay of features that necessitate the use of more powerful learners for flight delay prediction.

\section{Methodology}\label{chapter:methodology}
\begin{figure*}
\begin{center}
    \includegraphics[width=0.9\textwidth]{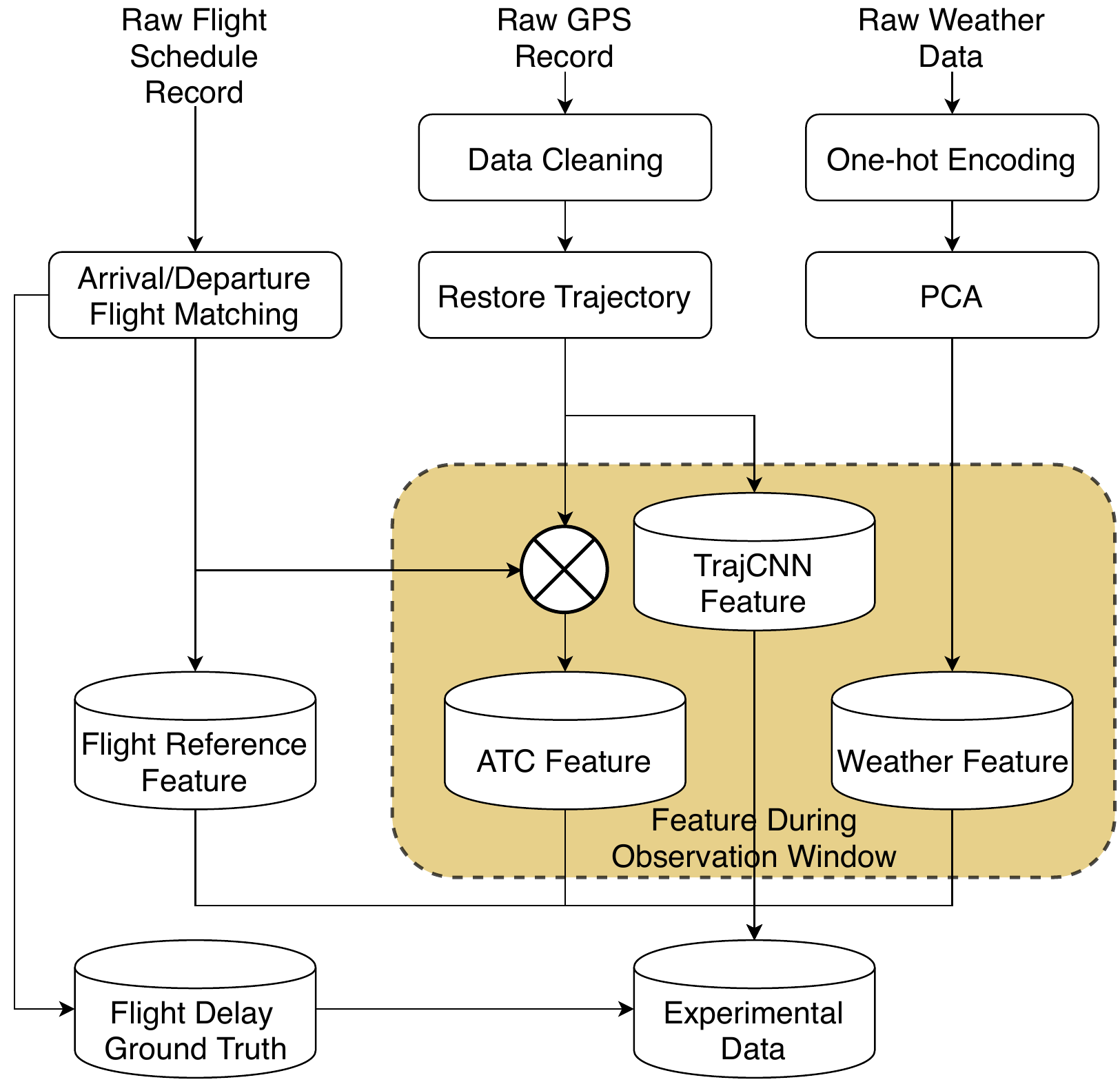}
\caption{The demonstration of the overall system architecture.} \label{fig_system}
\end{center}
\end{figure*}

The architecture of the flight departure delay time prediction system will be described in this section. As shown in Figure \ref{fig_system}, multiple datasets are gathered including: the GPS sensor data of aircraft and vehicles in the tarmac area; the historical flight data from major airline companies; and the associated weather data. Besides the noise found, some datasets had redundant portions, and some were incomplete. Therefore, a proper pre-processing procedure is required. To this end, we apply the methods mentioned in \cite{shao2019onlineairtrajclus} to make the datasets ready for the next stage. More specific details will be discussed in the following section \ref{sec_Data_cleaning}.

In the feature extraction stage, various methods were used to extract features from the cleaned datasets. First, a partition of the whole ground surface of LAX is used to separate different areas apart. Trajectories and GPS dots are counted to represent the traffic density of those areas to form the baseline ATC, whereas a more granular 2D histogram method was used to construct TrajCNN Features. A principal component analysis (PCA) is used to reduce the dimensionality of the weather data for the non-deep learning methods. The deep learning methods could extract latent features directly from the raw data. We also extracted multiple features from the combination of the historical flight scheduling data and trajectory sensor data.

\subsection{Data Pre-processing}\label{sec_Data_cleaning}
\begin{figure*}[htbp]
\centering
\subfigure[Three types of Tarmac area]{
\includegraphics[width=0.45\linewidth]{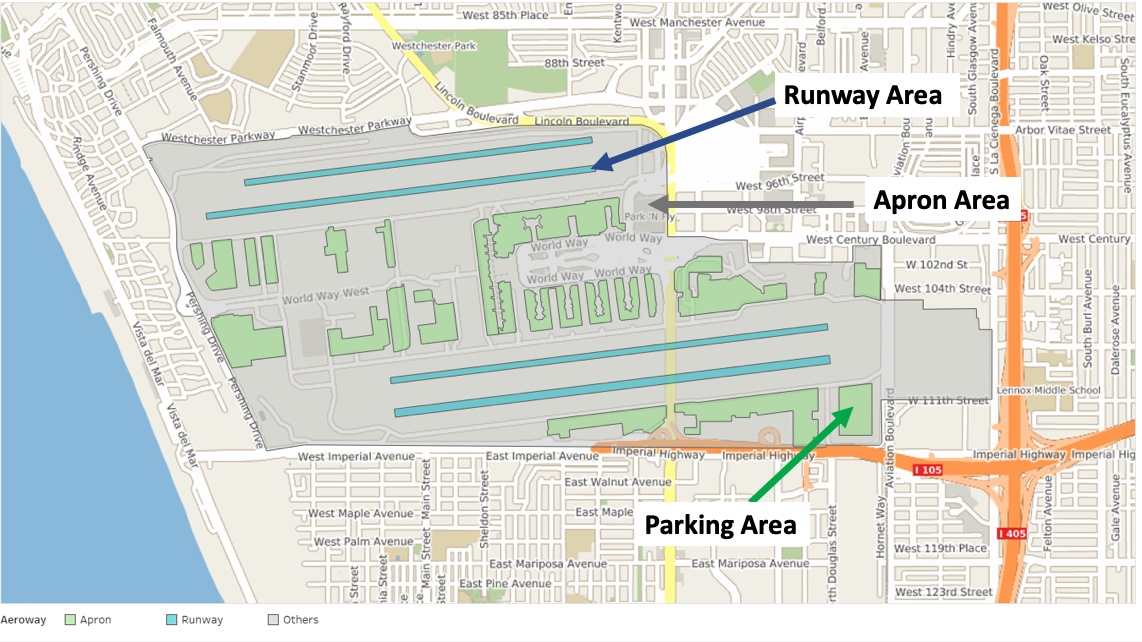}
}
\subfigure[Heatmap of the aircraft position]{
\includegraphics[width=0.45\linewidth]{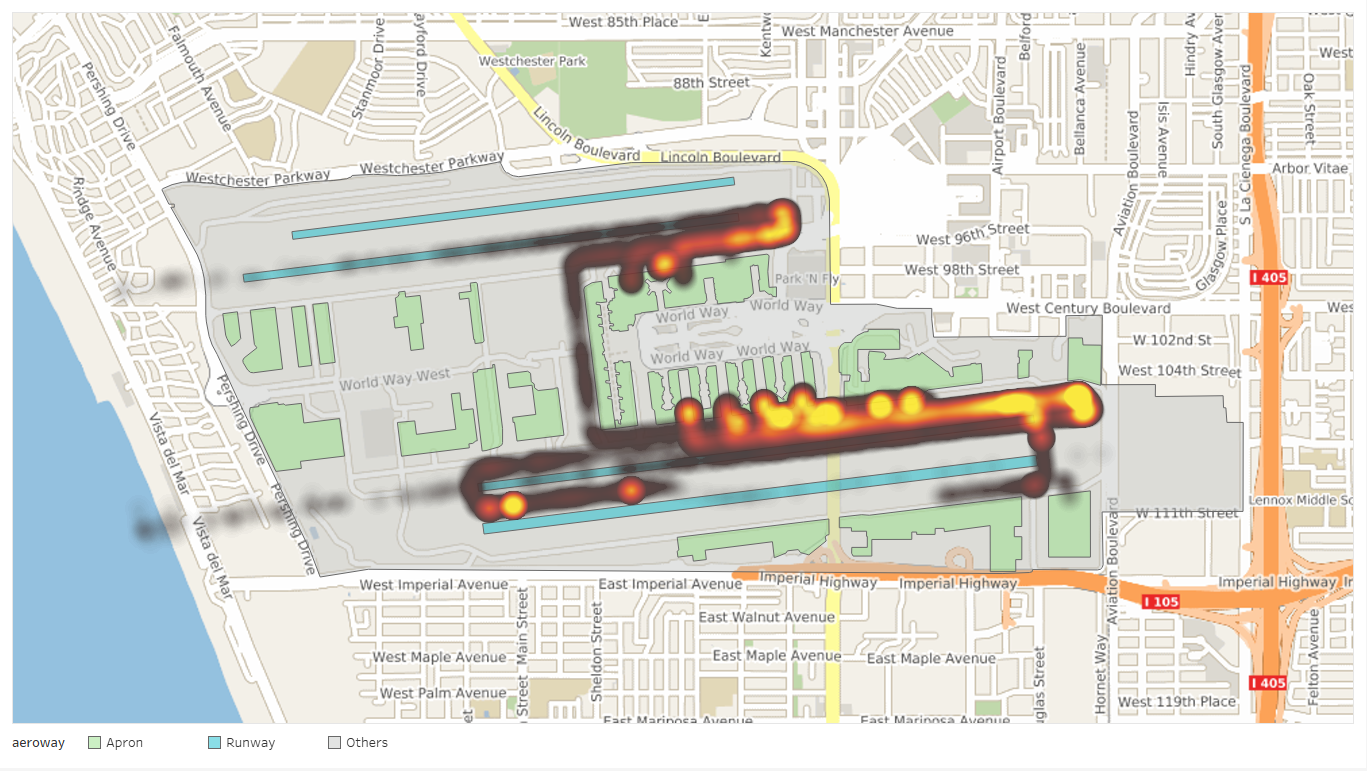}
}
\caption{The tarmac area is classified into 3 types: apron/runway/parking area.}
\label{fig_tarmacmap}
\end{figure*}

As mentioned in the previous section, some datasets contain irrelevant and redundant data. For example, the raw trajectory sensor data comprises the GPS records for ground vehicles, aircraft, and the historical flight scheduling data contains ICAO (International Civil Aviation Organization) number. Besides filtering out the irrelevant data points, it is also required to restore a continuous trajectory of each flight based on the noisy data, which includes some random jitter that falls within some impossible area. Therefore, we developed a data pipeline that can ignore these random jitters. We also calculated the average speed of the aircraft using the distance delta between two GPS records and eliminated the outliers. After that, the pipeline could form a trajectory of each flight based on its call sign and date since the same call sign is shared by all the flights travelling on the same route. At the end of this stage, each trajectory will be labelled into three different types based on which tarmac area the trajectory encountered. An illustration of the three tarmac types: parking area, apron area, and runway area are shown in Figure \ref{fig_tarmacmap}. Here the parking area indicates that most aircraft move slowly in this area. Indeed, we checked the map of the airport and found this area is for cargo purpose.

\subsection{Feature Extraction}
The features we used in this study can be classified into three categories: Weather Features, ATC features and TrajCNN Features. The first two are defined in Section \ref{chapter:definition}; more details about the extraction of TrajCNN Feature are described in the following section.

We matched every departure flight in LAX airport with their corresponding arrival flight record in the historical scheduling table data. That will help to introduce the possible delay propagation effect into consideration since the delay of the current flight might not just come from the current situation, but also has some relationship with the status of its previous arrival flight. 

For the ATC data, we extracted features for each flight during its observation window. As shown in Figure \ref{fig_flight}, the observation window is a duration period before the prediction time. We extracted features from the ATC data listed in Table \ref{tab:atc_desciption}. These features comprise a representation of the traffic density across all three aforementioned tarmac area areas. The number of potential landing/take-off aircraft is also calculated to help reveal the complexity.

\begin{table}[htbp]
\caption{The referenced weather features used in the feature vectors.}\label{tab:weather_vectors}
\centering
\begin{tabular}{l||l|l}
\hline 
\hline 
\textbf{Attribute Name} & \textbf{Description} & \textbf{Type} \\
\hline
Temperature & \tabincell{l}{The actual temperature at LAX (Fahrenheit).} & Real value\\
\hline
Dew point & \tabincell{l}{The temperature of dew point.} & Real value\\
\hline
Humidity & \tabincell{l}{Humidity at that time (\%).} & Real value\\
\hline
Wind direction & \tabincell{l}{Wind Direction represented by string (e.g. SE).} & Categorical\\
\hline
Wind speed & \tabincell{l}{The speed of the wind in that certain direction \\(mph).} & Real value\\
\hline
Wind gust & \tabincell{l}{The potential increase in the speed of the wind.} & Real value\\
\hline
Pressure & \tabincell{l}{The ambient air pressure at LAX.} & Real value\\
\hline
Condition & \tabincell{l}{Describing the state of the atmosphere in terms \\of temperature and wind and clouds and \\precipitation (e.g. Cloudy).} & Categorical\\
\hline
\hline 
\end{tabular}
\end{table}

For the non-deep learning methods, we performed a principal component analysis (PCA) on the weather data to reduce the dimensionality. The PCA is a statistical procedure which can reduce the dimension of a feature set while still retaining most of the knowledge in it by transferring them into another set of variables that are linearly uncorrelated \cite{li2014principal}. After this process, the weather data in Table \ref{tab:weather_vectors} manages to get 18 principal components while still containing its information. Furthermore, we use the most recent weather data within or near the observation window for each flight.

Graph embedding is widely used in deep learning models especially for trajectory data \cite{chen2019trip2vec, wang2018efficient}. However, for our model, we use the raw data because we compared the results between using raw data and the graph embedding results and found the raw data is a better option. It is because it is difficult for extract accurate graph network from airport.

We converted the Wind Direction attribute, which was initially a categorical attribute (e.g. N/S/W/E/NW, etc.), to bearings in radians. We have applied multiple machine learning approach to estimate the departure delay with different data. We found that weather data, in most experiments, shows no contributions to the prediction accuracy. Therefore, we did not consider it in our deep learning model.

\begin{table*}[htbp]
\caption{The ATC feature description}\label{tab:atc_desciption}
\centering
\begin{tabular}{l||l}
\hline 
\hline 
 \textbf{Attribute Name} & \textbf{ Description} \\
\hline
 takeoff\_plan & \tabincell{l}{The number of aircrafts planning to take-off \\in the Delay-Prediction Time Gap.} \\
 \hline
 takeoff\_num & \tabincell{l}{The number of aircrafts that actually take-off \\during the Observation Window. }\\
 \hline
 landing\_plan & \tabincell{l}{The number of aircrafts planing to land in the \\Delay-Prediction Time Gap. }\\
 \hline
 landing\_num & \tabincell{l}{The number of aircrafts that actually land \\during the Observation Window. }\\
 \hline
 apron\_point & \tabincell{l}{The discrete aircraft-GPS point count in the \\apron area. }\\
 \hline
 runway\_point & \tabincell{l}{The discrete aircraft-GPS point count in the \\runway area.  }\\
 \hline
 patrol\_point & \tabincell{l}{The discrete aircraft-GPS point count in the \\patrolling area. }\\
 \hline
 apron\_traj & \tabincell{l}{The discrete aircraft-trajectory count in the \\apron area. }\\
 \hline
 runway\_traj & \tabincell{l}{The discrete aircraft-trajectory count in the \\runway area. }\\
 \hline
 patrol\_traj & \tabincell{l}{The discrete aircraft-trajectory count in the \\patrolling area. }\\
 \hline
 \hline 
\end{tabular}
\end{table*}

\subsection{TrajCNN Features} \label{sec_trajcnn_features}
After we preprocess the GPS dataset, we then used these to construct TrajCNN features to capture the spatio-temporal information from the airport tarmac for our deep learning model. For every flight, $\mathbf{f}$, we first selected a subset of GPS points that is within the flight's observation window, $\mathbf{P}'$. Then, we divided the airport into a $28 \times 28$ grid and perform three 2D histograms. We chose the grid size of 28 as it is one of the common resolutions for CNN, such as MNIST. The first channel is the counting channel while the subsequent two channels are the of sum x- and y- velocity component on each grid, respectively. Then we use a global scalar scaler to scale the pixel values between 0 and 1. We stack these three channels into one image. This image is the TrajCNN features associated with flight $\mathbf{f}$.

\subsection{SVR}
Support Vector Machine (SVM) is a technique that was developed by Vladimir Vapnik in 1995 \cite{smola2004tutorial}. It constructs a hyperplane in a high-dimensional feature space. It was proposed to get a better result for classification problems. Later, Drucker et al. \cite{drucker1997support} made some modifications to make it suitable for regression tasks. This newer version is known as Support Vector Machine. The most crucial step in this method is the production of the hyperplane. Exploiting kernel methods to project the features into higher dimensional space could make the classes linearly separable and improve results. There are two major factors in this model: the parameter, $\varepsilon$, affects how close the fitting can be, and the parameter, $C$, controls model's patience on the error. Changes in $C$ causes the degradation of the model's generalisation ability.

\subsection{MLP}
The Multilayer Perceptron (MLP) is a feed-forward neural network, and its structure is very self-explanatory. Each layer contains several units, and there are one or more hidden layers that sit between the input and the output layer \cite{haykin1994neural}. An activation function can be applied to each node in this model, except the ones on the input layer. By introducing the activation function,  the performance on fitting non-linear relationships can be improved. In addition, increasing the number of hidden layers can also help on the same goal, but might also cause over-fitting. In our experiments, the MLP model has two hidden layers.

\subsection{LightGBM}
Gradient Boosting Decision Tree (GBDT) is a popular machine learning algorithm in recent years, and it is an effective way of building a predictive model.  The basic idea is to generate multiple weak learners and use an additive model on those weak learners to minimise the objective function, which is typically formed as a loss function. In this paper, we use a newer variant of it, LightGBM, which was recently developed by Microsoft \cite{ke2017lightgbm} as the GBDT framework for the experiment. Although there are other implementations such as XGBoost and pGBRT, the reason why we choose LightGBM is due to its several improved features:

\begin{itemize}
    \item The usage of histogram-based algorithms suppress the need of going through all discrete values. That ends up in faster training speed and lower memory consumption.
    \item The implementation some advanced network communication algorithms to utilise the multi-core processors for parallel learning.
\end{itemize}

\begin{figure}
\includegraphics[width=\linewidth]{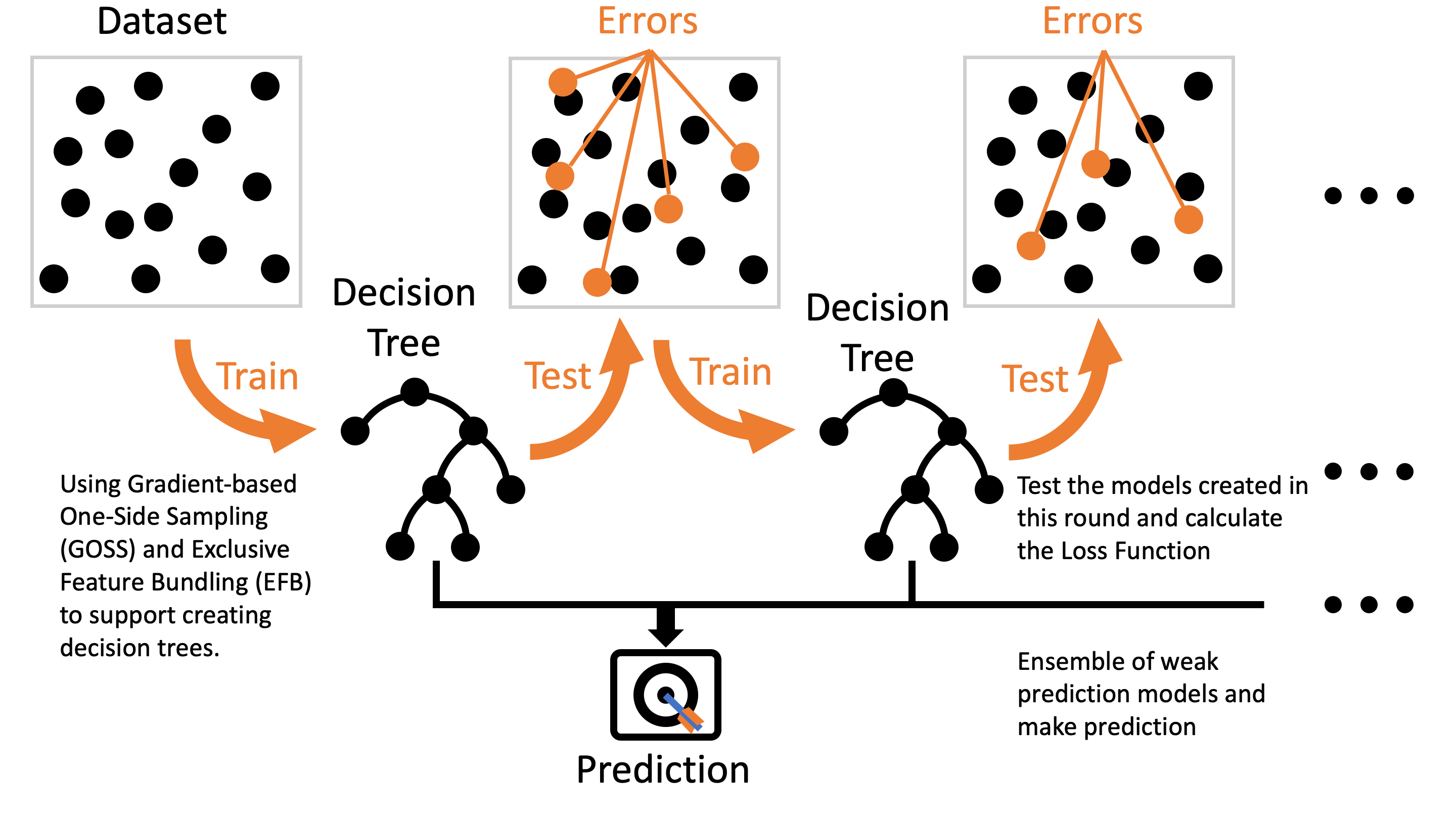}
\caption{The illustration of the LightGBM model creation process} \label{fig_lightGBM}
\end{figure}

As shown in Figure \ref{fig_lightGBM}, the primary process in the LightGBM training is to repeatedly create estimator, which is a weaker decision tree and test the remaining loss based on all the weaker trees created so far. 

Since departure delay prediction is a regression task, we use the Root Mean Squared Error (RMSE) as the loss function. 

\subsection{TrajCNN}

\begin{figure*}
\includegraphics[width=\linewidth]{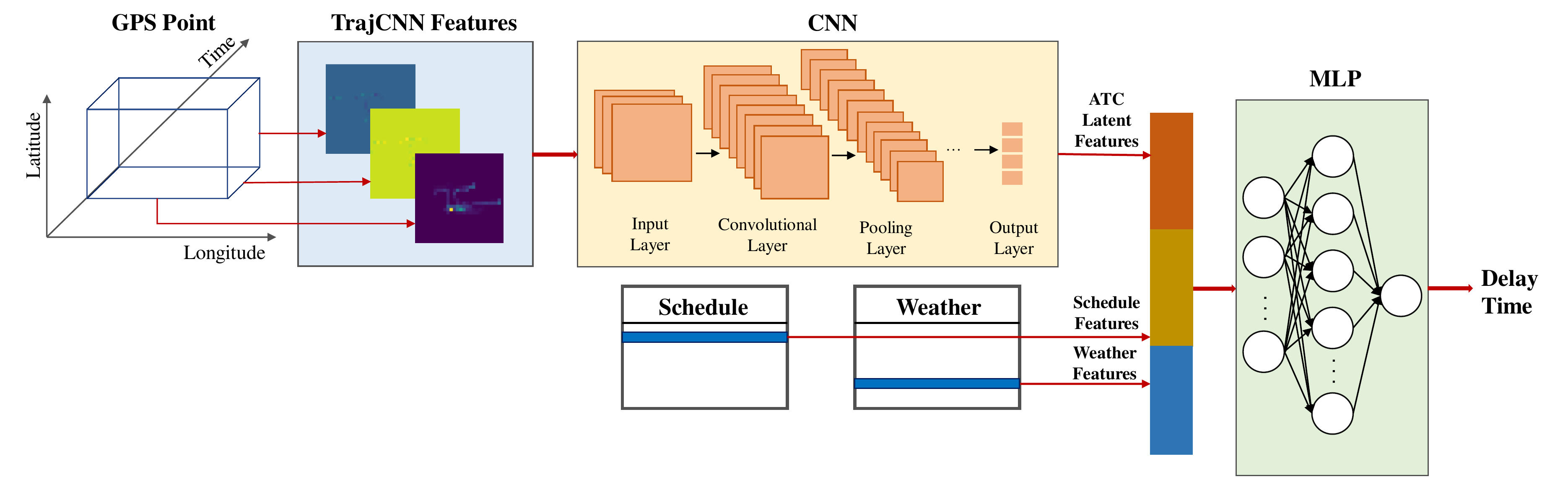}
\caption{TrajCNN architecture. We develop TrajCNN features to capture the spatiotemporal information from the GPS points on the airport tarmac. We use CNN to map these features into Air Traffic Complexity (ATC) latent features to represent the delay related complexity of the airport. Then, we fuse this with the schedule information before feeding it to an MLP regressor.} \label{fig:trajcnn_architecture}
\end{figure*}

The Convolutional Neural Network (CNN) is one of the first architecture in the now ubiquitous deep learning paradigm \cite{lecun1989backpropagation} \cite{le1989handwritten}.
It has been shown to work well with image and image-like data \cite{simonyan2014very} \cite{he2016deep} \cite{krizhevsky2012imagenet}.
As a result, many have attempted to engineer image-like features out of various problems \cite{tas2018cnn}.
This includes spatio-temporal predictions such as traffic predictions
\cite{yao2019revisiting} \cite{wang2016traffic} \cite{ma2017learning}.

Airport tarmac contains spatio-temporal information that would aid in solving the flight delay prediction problem \cite{shao2019flight}. Unlike the previous work that divides the airport into tailored areas, we construct a novel TrajCNN feature that would capture more information at a higher granularity as we make less amounts of aggregation. The deep learning architecture would pick up latent spatial-temporal features that might be lost during the coarse level aggregation of the previous work.

The CNN architecture is as follows. The TrajCNN features are firstly fed into two blocks of convolutional blocks. Each block consists of a convolutional layer, a ReLu activation layer, and a 2 $\times$ 2 max-pooling layer, and the convolutional layer has only one 3 $\times$ 3 filter. After flattening, it outputs the ATC latent features that capture the delay related complexities of the airport. We fuse this with the remaining flight features, $\mathbf{f_{x})}$ and use it as an input to a fully connected multilayer perceptron (MLP) with the delays $\mathbf{f_{y}}$ as the output. The detail of our architecture can be shown in Figure \ref{fig:trajcnn_architecture}.

\section{Experiment and Results} \label{chapter:ExperimentAndResults}
\subsection{Datasets}
In this section, we describe all data we used for our prediction in details.
\subsubsection{Schedule Table Data}
The first dataset consists of reference data operated by the Bureau of Transportation Statistics of the United State Department of Transportation \cite{historicaldata}. This dataset is an open dataset, and we extracted the relevant parts from it which comprise all flight records from the 1\textsuperscript{st} of July, to the 31\textsuperscript{st} of August. Both arrival and departure flight information occurred in LAX are included since we can extract features about the delay propagation effect from the former while the latter provides the ground truth of the departure delay that we tried to predict. Except for the overall flight delay (both arrival and departure), the delay is partitioned and categorised into five different types for various causes. Additionally, it also contains the Call-sign/Tail-number, which helps us to match it with our GPS trajectories data.

\subsubsection{Weather Data}
The second dataset in this experiment is reference data collected from website \cite{weathersource}. This weather data mainly covers the Los Angeles Airport (LAX) and some surrounding regions. It is gathered hourly, which provides a fine-grade tracking of the weather condition. The essential numerical-based features are temperature and humidity level. Meanwhile, the categorical features, Condition and Wind Direction, are critical for later stages. We apply one-hot encoding on the categorical features. More details are shown in Table \ref{tab:feature_vectors}.


\subsubsection{Airport GPS Trajectories Sensor Data}
The third reference dataset is a private dataset which consists of GPS observation of all vehicles and aircraft in the LAX (Los Angeles Airport). It is collected by the United States' Federal Aviation Administration's (FAA's) System Wide Information Management (SWIM) program. Trajectories of all vehicles can be restored based on this dataset. In the spatial domain, it covers all the tarmac area of LAX while the vertical altitude goes up to 15000 metres. In the temporal domain, the data contains seven weeks of data from the 1\textsuperscript{st} of July, 2016, to the 18\textsuperscript{th} of August, 2016. There are around 11 million GPS points fall within this range, which include both ground vehicles and aircraft locations. We managed to restore 43,503 trajectories from it, which belongs 6,518 vehicles. 

\subsection{Experimental Setup}
We withhold the last (28.6\%, 2 weeks) of the dataset for testing, while we use the remaining (71.4\%, 5 weeks) for training and validation. We performed cross-validation by dividing the dataset into 5 folds. The reason we chose this method of cross-validation, instead of using the random split method, is because each data point is temporally correlated with the other. We do this to ensure no temporal correlation among each fold. 

\subsubsection{MLP}

The hyper-parameter search space for MLP consist of:
\begin{itemize}
    \item Number of nodes sampled from the following probability distribution: $\mathcal{N}(x)=\{nint(N\times2^{x})|x$ is distributed randomly over $2\leq x\leq 12\}$ where $N$ is a normalisation constant
    \item Number of layers sampled from the following probability distribution: $\mathcal{L}(x)=\{nint(R\times x)|x$ is distributed randomly over $1\leq x\leq 11\}$ where $L$ is a normalisation constant
    \item Adam learning rate is sampled from the following probability distribution: $\mathcal{R}(x)=\{R\times2^{-x}|x$ is distributed randomly over $9\leq 17\leq 2\}$ where $R$ is a normalisation constant
\end{itemize}

Where $nint(x)$ is a nearest integer rounding function.



\subsubsection{LightGBM}
The hyper-parameters for LightGBM in this experiment are shown in Table \ref{tab:hyperparameter_attributes}, which also summarise the hyper-parameters for other models as well. The three most important parameters are: learning rate, number of estimators, and number of leaves. We also applied the early stop mechanism to avoid the overfitting. All other parameters were left at their default settings. All hyper-parameters are held unchanged through the whole testing procedure, which includes a 5-fold cross-validation experiment and its following application on the testing data-set.

\subsubsection{TrajCNN}
Since the training process for deep learning models is time-consuming, we did not perform 5-fold validation. Instead, we randomly split the data with the ratio of 7-1-2 for training-validation-testing, respectively. Our hyper-parameter search space is as follows:

\begin{itemize}
    \item The number of fully connected layers is sampled from the following probability distribution: 
    $\mathcal{L}(l)=\{nint(L\times l)|l$
    is distributed randomly over
    $1\leq l\leq 2\}$ where $L$ is a normalisation constant.
    
    \item The number of nodes in the fully connected layer is sampled from the following probability distribution: 
    $\mathcal{N}(n)=\{nint(N\times2^{n})|n$
    is distributed randomly over
    $1\leq n\leq 11\}$ where $N$ is a normalisation constant.
    
    \item The number of convolutional layers is sampled from the following probability distribution: 
    $\mathcal{C}(c)=\{nint(C\times c)|c$
    is distributed randomly over
    $1\leq c\leq 4\}$ where $C$ is a normalisation constant.
    
    \item The number of filter in convolutional layer is sampled from the following probability distribution: 
    $\mathcal{F}(f)=\{nint(F\times2^{f})|f$
    is distributed randomly over
    $1\leq f\leq 6\}$ where $F$ is a normalisation constant.
    
    \item The batch size is sampled from the following probability distribution: 
    $\mathcal{B}(b)=\{nint(N\times2^{b})|b$
    is distributed randomly over
    $1\leq b\leq 8\}$ where $B$ is a normalisation constant.
    
    \item The Learning rate is sampled from the following probability distribution: 
    $\mathcal{R}(r)=\{N\times10^{-r}|r$
    is distributed randomly over
    $3\leq r\leq 6\}$ where $R$ is a normalisation constant.
\end{itemize}

We performed around 100 experiments and chose the hyper-parameter set with the best validation RMSE. The best hyper-parameters can be found in table \ref{tab:hyperparameter_attributes}. We then use this hyper-parameter set to train the final model on both the training and validation dataset. Finally, we evaluate the final model on the test dataset.

\begin{table*}[!ht]
\caption{The Hyper-parameter using in this experiment}\label{tab:hyperparameter_attributes}
\centering
\begin{tabular}{l||p{0.20\textwidth}<{\centering}p{0.20\textwidth}<{\centering}}
\hline 
\hline 
Hyper-parameter & Attribute Name & Value \\
\hline 
\hline
 Model & \multicolumn{2}{c}{SVR}\\
\hline
 Kernel type & kernel & rbf \\
 The error term & C & 10000 \\
\hline 
\hline
 Model & \multicolumn{2}{c}{MLP}\\
\hline 
 Dropout & is\_dropout & False \\
 Num of epoch & n\_epoch & 3000 \\
 Early stopping round & n\_patience & 50 \\
 Num of layers & n\_layer & 2 \\
 Num of node & n\_node & 1553 \\
 Adam learning rate & adam\_learning\_rate & 0.00976563 \\
\hline 
\hline
 Model & \multicolumn{2}{c}{LigtGBM}\\
\hline 
 Learning rate & learning\_rate & 0.01 \\
 Number of estimators & n\_estimators & 16000 \\
 Number of leaves & num\_leaves & 39 \\
 \hline
 \hline
Model & \multicolumn{2}{c}{TrajCNN}\\
\hline 
Num of fully connected layer			&	n\_fc\_layer					&	1 \\
Num of node in fully connected layer 	&	n\_fc 						&	429 \\
Num of convolutional layer				&	n\_conv\_layer				&	2 \\
Num of filter in convolutional layer 	&	n\_conv 						&	1 \\
Batch Size								&	batch\_size					&	15 \\
Num of epoch 							&	n\_epoch 					&	200 \\
Early stop patience 					&	n\_early\_stop\_patience		&	20 \\
Optimizer								&	optimizer 					&	Adam \\
Learning rate							&	adam\_lr						&	3.48981e-05 \\
 \hline
 \hline
\end{tabular}
\end{table*}

\subsection{Experimental Results} \label{sec_exp_res}

In this section, we conducted three sets of experiments. In this first experiment, we evaluated the prediction results using different combinations of data sources and machine learning models. In the second experiment, we tested the temporal sensitivity of the learning model and validated the robustness of the model in flight delay time prediction. In the last set of experiments, we compared the importance of different features that we extracted and created from the historical, ATC, and weather datasets. We conducted these three sets of experiments to evaluate the performance of our proposed flight delay prediction framework and the importance of our proposed ATC features.


In the first set of experiments, we apply four conventional machine learning regressors: Linear Regression (LR), Support Vector Regressor (SVR), Multilayer Perceptron (MLP), LightGBM, and our proposed deep learning architecture, TrajCNN, to predict the flight delay using different combination of data sources (Historical, weather, and GPS data). In this experiment, we use the traditional regression evaluation metrics: Root Mean Square Error (RMSE) to measure the performance of flight delay time prediction using different models and combinations of data sources. A Lower RMSE shows less error between prediction value and ground truth.

As shown in Table \ref{tab:result_barchart}, LightGBM has the best overall performance. We can found out that there is a clear performance boost after adding spatiotemporal information, either in the form of ATC features, or TrajCNN features. In fact, without the spatiotemporal information, the RMSE of most algorithms is only around the standard deviation of the label. 

The best results come from the experiments that combines flight reference data, ATC features, and weather data. These results indicate that the ATC features play a significant role in predicting flight departure delay. 

Linear Regression performs poorly no on linear features such as ATC features, which other algorithms able to exploit well. Moreover, since we linearly decorrelate the data using PCA,  linear regression is received a minor boost in improvement.

In the second set of experiments, we choose two temporal parameters to validate the robustness of the prediction result. We use only LightGBM in this experiment because it is one of the best performers with the shortest training time. The first parameter is the length of the observation window. As illustrated in Figure \ref{fig_flight}, the observation window is between the time point we predict the flight delay and the time point we start to collect ATC and weather data. The other parameter is the delay-predicting gap which starts from prediction time and gate-out time (flight close the gate). From intuition, Longer delay-predicting gap is likely to lead to worse results due to higher uncertainty. Meanwhile, shorter observation window means less training data is used in the model, which is likely to cause worse performance. 

\begin{figure}[htbp]
\centering
\includegraphics[width=1.0\linewidth]{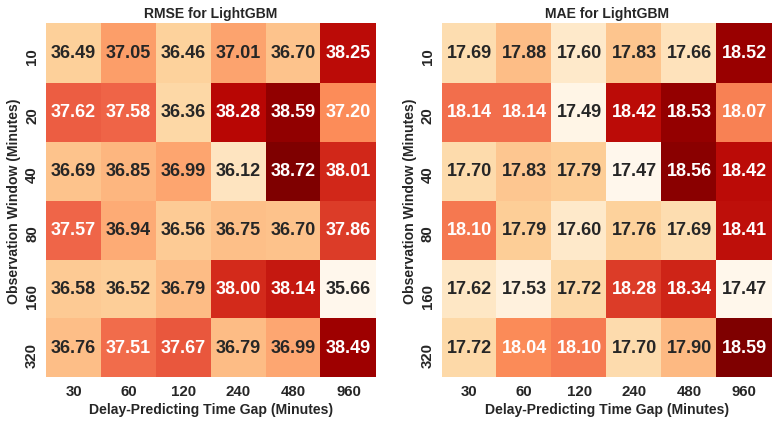}
\caption{The RMSE and MAE flight departure delay prediction result by LightGBM using the different observation window (x) and delay-prediction time gap (y) combinations.} \label{fig_confusion_matrix}
\end{figure}

\begin{table}[htbp]
\caption{The RMSE result for different models and different combination of different data sources. (* Not Applicable as TrajCNN have to use images from TrajCNN features to work.)}\label{tab:result_barchart}
\begin{adjustwidth}{-1cm}{}
\centering
\begin{tabular}{l||c|c|c|c|c}
\hline \hline
RMSE                                   & LR  & MLP                & SVR               & LightGBM           & TrajCNN           \\
\hline
Flight Reference Only                  & 44.3648          & 44.5027          & 44.3945         & 40.1237          & *                 \\
Reference+Weather                    & \textbf{44.3060} & 45.6198          & 44.0933         & 39.2137          & *                 \\
Reference+ATC                        & 44.3553          & 44.4964          & 43.9669         & 38.2711          & *                 \\
Reference+Weather+ATC               & 44.3117          & \textbf{44.0964} & \textbf{43.896} & \textbf{37.9478} & *                 \\
Reference+TrajCNN Features           & *                  & *                  & *                 & *                  & \textbf{37.2533}  \\
\tabincell{l}{Reference+Weather\\ \tab+TrajCNN Features} & *                  & *                  & *                 & *                  & 38.6123         \\
\hline \hline
\end{tabular}
\end{adjustwidth}
\end{table}


In summary, prediction accuracy is stable with even less training data. Furthermore, even the prediction time is four hours ahead of the flight delay event, the accuracy (around 17 minutes) is acceptable.

In the third experiment, we compare the importance of features we used in ATC dataset, weather condition dataset and historical dataset. We use the parameter $feature\_ importance$ recorded by the LightGBM regressor, which captures the numbers of times a specific feature is used to `split' a tree \cite{featureimportance}. The higher times this value is, the more information this feature provides, which helps the model to differentiate the situation.

\begin{figure*}
\includegraphics[width=\textwidth]{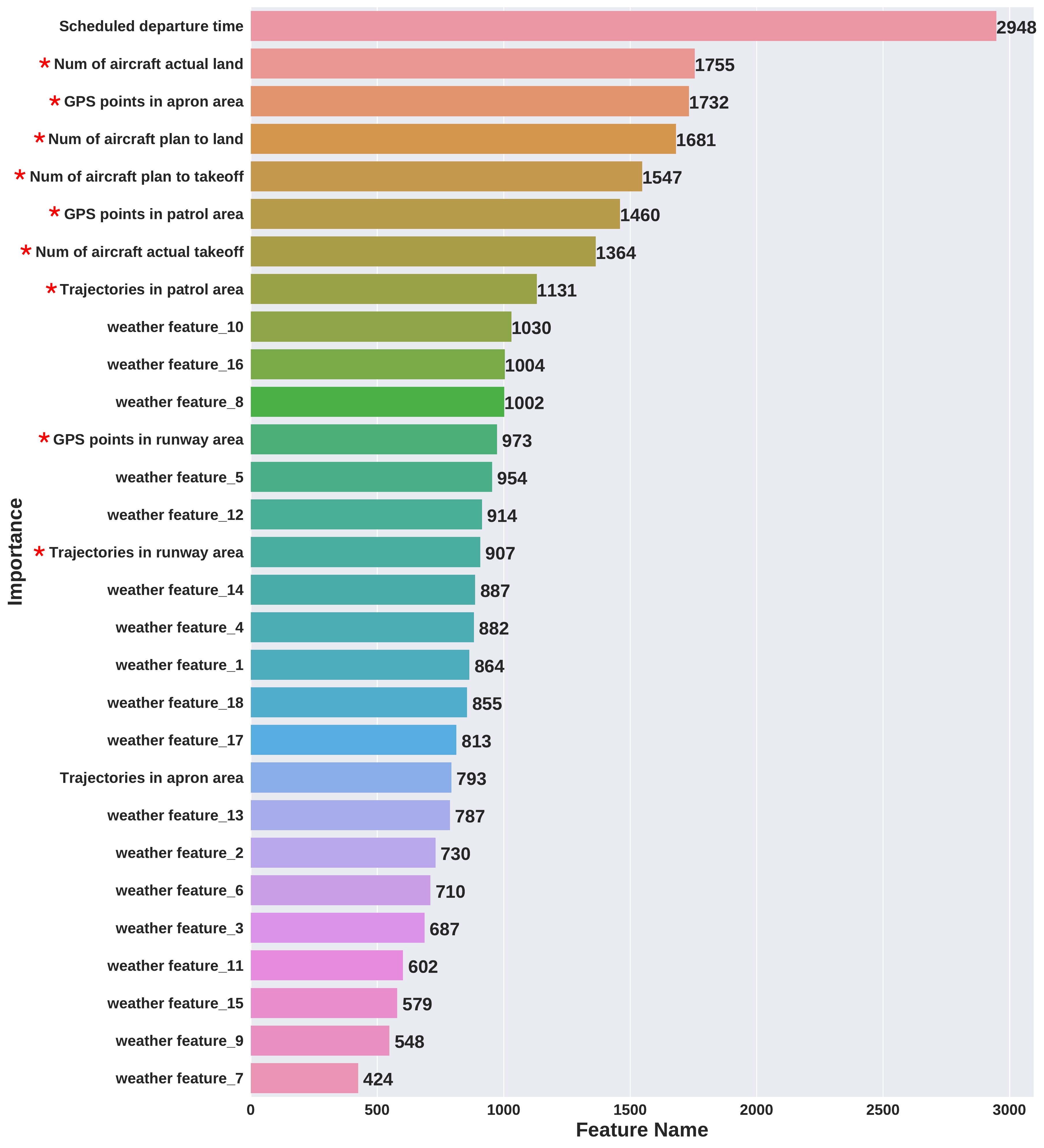}
\caption{The feature importance for Weather and ATC features. All ATC features in this graph has a red asterisk before them.} \label{fig_feature_importance}
\end{figure*}

Figure \ref{fig_feature_importance} shows the feature importance after the training process. Firstly, the historical data is still the most important feature (Schedule Departure Time) in predicting departure delay. Secondly, we can find that most of our selected ATC features have significantly higher importance than the weather features, which also supports the result in the first experiment that the ATC features does contribute more in this task.

\section{Discussion and Future Work}\label{chapter:Discussion}
Our proposed features and framework achieve an acceptable prediction performance in flight departure delay problem. However, there remain many limitations and potential work for the future. Firstly, this work mainly focuses on predicting the departure delay using airport traffic data. We did not consider many other traditional factors such as air traffic or airport traffic control system. Secondly, due to the difficulty to collect data from airports, we only chose the data from one airport, which limits the generalisation of our proposed methods. Although our proposed solutions do not rely on any specific factor of this airport, various data are still needed. Additionally, the seasonal pattern also plays an important role in flight delay as well. However, due to the range of the data provided by FAA, we only evaluate our system using a few months data. We plan to apply our proposed framework to different airports in a longer time in future work. Thirdly, the proposed deep learning networks can still be improved with more state-of-the-art techniques such as attention mechanism and assign the temporal prediction ability to this network using recurrent components. Fourthly, due to the purpose of this paper is to estimate the departure delay using only on-ground data at the airport, many factors such as conditions of air route, destination, and airline (Hub or Non-Hub) are not taken into account, which can also significantly affect the departure time. We plan to incorporate these factors with on-ground GPS data to achieve higher accuracy in the future. Additionally, many STOA CNN-based models have not been applied to this data since the rapid development of the deep learning area. Lastly, the existing work predicts the flight delay 4 hours before the scheduled departure time based on the observation window. However, sometimes, people or airport traffic controller needs a long horizontal prediction. We plan to apply the existing model to various needs and long-term prediction in the future.

\section{Conclusion}\label{chapter:conclusion}
We have proposed to use airport traffic complexity (ATC) data combined with traditional environmental factors to predict the flight departure delay from heterogeneous sensor data. We have designed, implemented, and evaluated an end-to-end deep learning-based framework to infer the delay time from a sequence of the snapshot of aircraft trajectories at the airport. We show in our evaluation that our approach allows for estimating flight departure delay with average deviations to the ground truth of fewer than 20 minutes.
http://www.cs.sjtu.edu.cn/~yaobin/ccf.htm
Also, we have demonstrated that the airport situational awareness map plays a more critical role in departure delay time prediction than weather conditions, and analyse the correlation between flight departure delay and ATC components. We also demonstrated that LightGBM outperforms other classic regressors using ATC feature, and a sequence of the snapshot of aircraft can be used to represent the aircraft trajectories and predict the flight delay. 

\section*{Acknowledgement}
This research is funded by Northrop Grumman Corporations USA for "Spatio-temporal Analytics for Situation Awareness in Airport Operations" project, RMIT University.
We would like to also acknowledge the support of the CSIRO Data61 Scholarship for Arian Prabowo and RMIT Research Stipend Scholarship (RRSS) for Sichen Zhao. We also acknowledge the support of ARC Discovery Project \textit{DP190101485}. 

\bibliography{mybibfile}

\begin{thebibliography}{10}
\expandafter\ifx\csname url\endcsname\relax
  \def\url#1{\texttt{#1}}\fi
\expandafter\ifx\csname urlprefix\endcsname\relax\def\urlprefix{URL }\fi
\expandafter\ifx\csname href\endcsname\relax
  \def\href#1#2{#2} \def\path#1{#1}\fi

\bibitem{GAO19}
{U.S. Government Accountability Office}, Airline consumer protections
  additional actions could enhance dot’s compliance and education efforts,
  Tech. rep., Department of Transportation (Nov 2018).

\bibitem{GAO11}
{U.S. Government Accountability Office}, Airline passenger protections more
  data and analysis needed to understand effects of flight delays, Tech. rep.,
  Department of Transportation (Sep 2011).

\bibitem{Kim2016}
Y.~J. Kim, S.~Choi, S.~Briceno, D.~Mavris, {A deep learning approach to flight
  delay prediction}, in: AIAA/IEEE Digital Avionics Systems Conference -
  Proceedings, Vol. 2016-December, 2016.
\newblock \href {http://dx.doi.org/10.1109/DASC.2016.7778092}
  {\path{doi:10.1109/DASC.2016.7778092}}.

\bibitem{Sternberg2017}
A.~Sternberg, J.~Soares, D.~Carvalho, E.~Ogasawara,
  \href{http://arxiv.org/abs/1703.06118}{{A Review on Flight Delay
  Prediction}}\href {http://arxiv.org/abs/1703.06118}
  {\path{arXiv:1703.06118}}.
\newline\urlprefix\url{http://arxiv.org/abs/1703.06118}

\bibitem{ahmadbeygi2010decreasing}
S.~Ahmadbeygi, A.~Cohn, M.~Lapp, Decreasing airline delay propagation by
  re-allocating scheduled slack, IIE transactions 42~(7) (2010) 478--489.

\bibitem{tu2008estimating}
Y.~Tu, M.~O. Ball, W.~S. Jank, Estimating flight departure delay
  distributions-a statistical approach with long-term trend and short-term
  pattern, Journal of the American Statistical Association 103~(481) (2008)
  112--125.

\bibitem{abdelghany2004model}
K.~F. Abdelghany, S.~S. Shah, S.~Raina, A.~F. Abdelghany, A model for
  projecting flight delays during irregular operation conditions, Journal of
  Air Transport Management 10~(6) (2004) 385--394.

\bibitem{rebollo2012characterization}
J.~J. Rebollo de~la Bandera, Characterization and prediction of air traffic
  delays, Ph.D. thesis, Massachusetts Institute of Technology (2012).

\bibitem{ahmadbeygi2008analysis}
S.~AhmadBeygi, A.~Cohn, Y.~Guan, P.~Belobaba, Analysis of the potential for
  delay propagation in passenger airline networks, Journal of air transport
  management 14~(5) (2008) 221--236.

\bibitem{mogford1995complexity}
R.~H. Mogford, J.~Guttman, S.~Morrow, P.~Kopardekar, The complexity construct
  in air traffic control: A review and synthesis of the literature., Tech.
  rep., CTA INC MCKEE CITY NJ (1995).

\bibitem{trub2018monitoring}
R.~Tr{\"u}b, D.~Moser, M.~Sch{\"a}fer, R.~Pinheiro, V.~Lenders, Monitoring
  meteorological parameters with crowdsourced air traffic control data, in:
  Proceedings of the 17th ACM/IEEE International Conference on Information
  Processing in Sensor Networks, IEEE Press, 2018, pp. 25--36.

\bibitem{prabowo2019coltrane}
A.~Prabowo, P.~Koniusz, W.~Shao, F.~D. Salim, Coltrane: Convolutional
  trajectory network for deep map inference, in: The 6th ACM International
  Conference on Systems for Energy-Efficient Buildings, Cities, and
  Transportation (BuildSys '19), November 13--14, 2019, New York, NY, USA,
  IEEE, 2019.

\bibitem{shao2019onlineairtrajclus}
W.~Shao, F.~D. Salim, J.~Chan, K.~Qin, J.~Ma, B.~Feest, Onlineairtrajclus: An
  online aircraft trajectory clustering for tarmac situation awareness, in:
  2019 IEEE International Conference on Pervasive Computing and Communications
  (PerCom, IEEE, 2019, pp. 192--201.

\bibitem{vouros2018big}
G.~A. Vouros, A.~Vlachou, G.~M. Santipantakis, C.~Doulkeridis, N.~Pelekis,
  H.~V. Georgiou, Y.~Theodoridis, K.~Patroumpas, E.~Alevizos, A.~Artikis,
  et~al., Big data analytics for time critical mobility forecasting: Recent
  progress and research challenges., in: EDBT, 2018, pp. 612--623.

\bibitem{andrienko2018creating}
G.~Andrienko, N.~Andrienko, Creating maps of artificial spaces to explore
  trajectories, in: Proceedings of the Workshop on Advanced Visual Interfaces
  AVI, ACM, 2018.

\bibitem{laudeman1998dynamic}
I.~V. Laudeman, S.~Shelden, R.~Branstrom, C.~Brasil, Dynamic density: An air
  traffic management metric, Tech. rep., NASA (1998).

\bibitem{kopardekar2007airspace}
P.~Kopardekar, A.~Schwartz, S.~Magyarits, J.~Rhodes, Airspace complexity
  measurement: An air traffic control simulation analysis, in: 7th USA/Europe
  Air Traffic Management R\&D Seminar, Barcelona, Spain, 2007.

\bibitem{djokic2010air}
J.~Djokic, B.~Lorenz, H.~Fricke, Air traffic control complexity as workload
  driver, Transportation research part C: emerging technologies 18~(6) (2010)
  930--936.

\bibitem{delahaye2003air}
D.~Delahaye, S.~Puechmorel, J.~Hansman, J.~Histon, Air traffic complexity based
  on non linear dynamical systems, 2003.

\bibitem{koros2003complexity}
A.~Koros, P.~S. Rocco, G.~Panjwani, V.~Ingurgio, J.-F. D'Arcy, Complexity in
  air traffic control towers: A field study. part 1. complexity factors, Tech.
  rep., FEDERAL AVIATION ADMINISTRATION TECHNICAL CENTER ATLANTIC CITY NJ
  (2003).

\bibitem{simic2015airport}
T.~K. Simi{\'c}, O.~Babi{\'c}, Airport traffic complexity and environment
  efficiency metrics for evaluation of atm measures, Journal of Air Transport
  Management 42 (2015) 260--271.

\bibitem{rebollo2014characterization}
J.~J. Rebollo, H.~Balakrishnan, Characterization and prediction of air traffic
  delays, Transportation research part C: Emerging technologies 44 (2014)
  231--241.

\bibitem{xu2008multifactor}
N.~Xu, L.~Sherry, K.~Laskey, Multifactor model for predicting delays at us
  airports, Transportation Research Record: Journal of the Transportation
  Research Board 10~(2052) (2008) 62--71.

\bibitem{shao2019flight}
W.~Shao, A.~Prabowo, S.~Zhao, S.~Tan, P.~Koniusz, J.~Chan, X.~Hei, B.~Feest,
  F.~D. Salim, Flight delay prediction using airport situational awareness map,
  in: 27th ACM SIGSPATIAL International Conference on Advances in Geographic
  Information Systems (SIGSPATIAL '19), November 5--8, 2019, Chicago, IL, USA,
  IEEE, 2019.

\bibitem{li2014principal}
J.~Li, R.~R. Linear, Principal component analysis (2014).

\bibitem{chen2019trip2vec}
C.~Chen, C.~Liao, X.~Xie, Y.~Wang, J.~Zhao, Trip2vec: a deep embedding approach
  for clustering and profiling taxi trip purposes, Personal and Ubiquitous
  Computing 23~(1) (2019) 53--66.

\bibitem{wang2018efficient}
X.~Wang, C.~Chen, Y.~Min, J.~He, B.~Yang, Y.~Zhang, Efficient metropolitan
  traffic prediction based on graph recurrent neural network, arXiv preprint
  arXiv:1811.00740.

\bibitem{smola2004tutorial}
A.~J. Smola, B.~Sch{\"o}lkopf, A tutorial on support vector regression,
  Statistics and computing 14~(3) (2004) 199--222.

\bibitem{drucker1997support}
H.~Drucker, C.~J. Burges, L.~Kaufman, A.~J. Smola, V.~Vapnik, Support vector
  regression machines, in: Advances in neural information processing systems,
  1997, pp. 155--161.

\bibitem{haykin1994neural}
S.~Haykin, Neural networks: a comprehensive foundation, Prentice Hall PTR,
  1994.

\bibitem{ke2017lightgbm}
G.~Ke, Q.~Meng, T.~Finley, T.~Wang, W.~Chen, W.~Ma, Q.~Ye, T.-Y. Liu, Lightgbm:
  A highly efficient gradient boosting decision tree, in: Advances in Neural
  Information Processing Systems, 2017, pp. 3146--3154.

\bibitem{lecun1989backpropagation}
Y.~LeCun, B.~Boser, J.~S. Denker, D.~Henderson, R.~E. Howard, W.~Hubbard, L.~D.
  Jackel, Backpropagation applied to handwritten zip code recognition, Neural
  computation 1~(4) (1989) 541--551.

\bibitem{le1989handwritten}
Y.~Le~Cun, L.~D. Jackel, B.~Boser, J.~S. Denker, H.~P. Graf, I.~Guyon,
  D.~Henderson, R.~E. Howard, W.~Hubbard, Handwritten digit recognition:
  Applications of neural network chips and automatic learning, IEEE
  Communications Magazine 27~(11) (1989) 41--46.

\bibitem{simonyan2014very}
K.~Simonyan, A.~Zisserman, Very deep convolutional networks for large-scale
  image recognition, arXiv preprint arXiv:1409.1556.

\bibitem{he2016deep}
K.~He, X.~Zhang, S.~Ren, J.~Sun, Deep residual learning for image recognition,
  in: Proceedings of the IEEE conference on computer vision and pattern
  recognition, 2016, pp. 770--778.

\bibitem{krizhevsky2012imagenet}
A.~Krizhevsky, I.~Sutskever, G.~E. Hinton, Imagenet classification with deep
  convolutional neural networks, in: Advances in neural information processing
  systems, 2012, pp. 1097--1105.

\bibitem{tas2018cnn}
Y.~Tas, P.~Koniusz, Deep residual learning for image recognition, in: BMCV,
  2018.

\bibitem{yao2019revisiting}
H.~Yao, X.~Tang, H.~Wei, G.~Zheng, Z.~Li, Revisiting spatial-temporal
  similarity: A deep learning framework for traffic prediction, in: AAAI
  Conference on Artificial Intelligence, 2019.

\bibitem{wang2016traffic}
J.~Wang, Q.~Gu, J.~Wu, G.~Liu, Z.~Xiong, Traffic speed prediction and
  congestion source exploration: A deep learning method, in: 2016 IEEE 16th
  International Conference on Data Mining (ICDM), IEEE, 2016, pp. 499--508.

\bibitem{ma2017learning}
X.~Ma, Z.~Dai, Z.~He, J.~Ma, Y.~Wang, Y.~Wang, Learning traffic as images: a
  deep convolutional neural network for large-scale transportation network
  speed prediction, Sensors 17~(4) (2017) 818.

\bibitem{historicaldata}
Weather-Underground, Los angeles international airport, california,
  \url{https://www.wunderground.com/history/daily/us/ca/los-angeles/KLAX/date/2016-7-31/},
  [Online; accessed 5-Apr-2019].

\bibitem{weathersource}
{Bureau of Transportation Statistics, U.S. Department of Transportation},
  Airline on-time statistics, \url{https://www.transtats.bts.gov/ONTIME/},
  [Online; accessed 5-Apr-2019] (2019).

\bibitem{featureimportance}
{Microsoft Corporation}, Lightgbm's python api,
  \url{https://lightgbm.readthedocs.io/en/latest/Python-API.html/}, [Online;
  accessed 5-Apr-2019].

\end{thebibliography}

\end{document}